%% file: main.tex
\documentclass[10pt,twocolumn,letterpaper]{article}

\usepackage{authblk}
\usepackage[pagenumbers]{capprobe}  % To force page numbers, e.g. for an arXiv version

\usepackage{capt-of}
\usepackage{xspace}
\usepackage[table]{xcolor}
\usepackage{enumitem}
\usepackage{amsmath,amsthm,amssymb,amsfonts,dsfont,pifont,bm,bbm,mathrsfs,mathtools,nicefrac,extarrows,relsize}
\usepackage{algorithm,algpseudocode,listings}
\usepackage{booktabs,multirow,adjustbox,diagbox,threeparttable,tabularray,setspace}
\usepackage[misc]{ifsym}
\usepackage{tablefootnote} % for table footnotes

\definecolor{capprobeblue}{rgb}{0.21,0.49,0.74}
\usepackage[pagebackref,breaklinks,colorlinks,citecolor=capprobeblue,bookmarks=false]{hyperref}
\usepackage{wrapfig}
\usepackage[capitalize]{cleveref}  % Should be loaded after 'hyperref', and works perfectly with 'subfigure'.
\newcommand\blfootnote[1]{%	
  \begingroup
  \renewcommand\thefootnote{}\footnote{#1}%
  \addtocounter{footnote}{-1}%
  \endgroup
}

\crefname{section}{Sec.}{Secs.}
\Crefname{section}{Section}{Sections}
\crefname{appendix}{App.}{Apps.}
\Crefname{appendix}{Appendix}{Appendices}
\crefname{table}{Tab.}{Tabs.}
\Crefname{table}{Table}{Tables}
\crefname{figure}{Fig.}{Figs.}
\Crefname{figure}{Figure}{Figures}
\crefname{equation}{Eq.}{Eqs.}
\Crefname{equation}{Equation}{Equations}
\crefname{theorem}{Thm.}{Thms.}
\Crefname{theorem}{Theorem}{Theorems}
\crefname{lemma}{Lem.}{Lems.}
\Crefname{lemma}{Lemma}{Lemmas}
\crefname{remark}{Rem.}{Rems.}
\Crefname{remark}{Remark}{Remarks}
\crefname{corollary}{Cor.}{Cors.}
\Crefname{corollary}{Corollary}{Corollaries}
\crefname{algorithm}{Alg.}{Algs.}
\Crefname{algorithm}{Algorithm}{Algorithms}
\definecolor{cellred}{RGB}{213, 123, 101}
\definecolor{cellgreen}{RGB}{0, 205, 0}
\definecolor{cellblue}{RGB}{54, 125, 189}
\definecolor{rankfirstbg}{RGB}{250, 205, 200}
\definecolor{ranksecondbg}{RGB}{205, 225, 245}
\definecolor{rankthirdbg}{RGB}{215, 240, 210}
\newcommand{\rankfirst}[1]{\cellcolor{rankfirstbg}\textbf{#1}}
\newcommand{\ranksecond}[1]{\cellcolor{ranksecondbg}\underline{#1}}
\newcommand{\rankthird}[1]{\cellcolor{rankthirdbg}\textit{#1}}
\newcommand{\rankmark}[2]{\begingroup\setlength{\fboxsep}{1.2pt}\colorbox{#1}{#2}\endgroup}
\newcommand{\rankmarkfirst}[1]{\rankmark{rankfirstbg}{\textbf{#1}}}
\newcommand{\rankmarksecond}[1]{\rankmark{ranksecondbg}{\underline{#1}}}
\newcommand{\rankmarkthird}[1]{\rankmark{rankthirdbg}{\textit{#1}}}
\definecolor{codegreen}{rgb}{0,0.6,0}
\definecolor{codegray}{rgb}{0.5,0.5,0.5}
\definecolor{codepurple}{rgb}{0.58,0,0.82}
\definecolor{backcolour}{rgb}{1.0,1.0,1.0}
\lstdefinestyle{mystyle}{
    backgroundcolor=\color{backcolour},
    commentstyle=\color{codegreen},
    keywordstyle=\color{magenta},
    numberstyle=\tiny\color{codegray},
    stringstyle=\color{codepurple},
    basicstyle=\ttfamily\scriptsize,
    breakatwhitespace=false,
    breaklines=true,
    captionpos=b,
    keepspaces=true,
    numbers=left,
    numbersep=5pt,
    showspaces=false,
    showstringspaces=false,
    showtabs=false,
    tabsize=2
}
\newcommand{\tocite}[1]{{\color{red} [TO CITE]}}
\newcommand{\methodname}{CapProbe}
\newcommand{\method}{\textit{\methodname}\xspace}
\title{CapProbe: Evaluating Detailed Image Captions via\\
Full-Scene Dense Question Answering}

\author{Mouxiao Huang\textsuperscript{$\dagger$}}
\author{Qiangyu Yan}
\author{Borui Jiang\textsuperscript{$\dagger$}}
\author{Han Shu\textsuperscript{\Letter}}
\affil{Huawei Technologies}

\begin{document}

\maketitle

\blfootnote{\small \textsuperscript{$\dagger$}~: This work was done when the authors were with Huawei Technologies. \Letter~: Corresponding author.}

\input{sec/0_abstract}
\input{sec/1_intro}
\input{sec/2_related_work}
\input{sec/3_method}
\input{sec/4_metrics}
\input{sec/5_exp}
\input{sec/6_conclusion}
\input{sec/7_ref}
\input{sec/8_appendix}

\end{document}

%% file: sec/0_abstract.tex
\begin{abstract}

Evaluating detailed image captions from Vision-Language Models (VLMs) requires going beyond surface-level semantic similarity.
Reference-based metrics (e.g., CIDEr, SPICE) and LLM-as-scorer protocols struggle to verify dense factual claims, while existing QA-based alternatives generally offer lower probe density, narrower domain coverage, or no explicit alignment between individual questions and segmented image regions.
We introduce \textbf{\method}, a full-scene dense QA benchmark that turns detailed caption evaluation into region-aligned factual checking.
Each image is decomposed into coarse semantic regions covering both foreground and background elements; for every retained region we generate multiple-choice questions spanning 10 semantic categories, forming a dense checklist of probed visual facts.
Guided by a two-tier taxonomy of 37 L1 domains and 219 L2 sub-domains, \method comprises 346 images, 1,868 regions, and 25,650 questions---averaging 74 QA pairs per image.
A language judge answers from the caption alone; an \emph{Uncertain} option and \emph{Effective Accuracy} provide a judge-dependent proxy for distinguishing unanswered probes from incorrectly resolved ones, and density-based metrics penalize verbose yet uninformative captions.
The protocol is cost-effective: by converting unconstrained scalar scoring into structured MCQ reading, it reduces open-ended scoring bias while remaining judge-conditioned, and yields relatively stable model rankings under a fixed reader.
Experiments on 13 VLMs show large Coverage gaps across models, a clear competency--efficiency trade-off, and failure modes that sparse or overlap-based evaluation often misses.
The benchmark data, annotations, and evaluation code will be released soon.

\end{abstract}

%% file: sec/1_intro.tex
\section{Introduction}\label{sec:intro}

\textbf{Detailed image captioning}---the task of generating exhaustive textual narratives 
that encapsulate objects, fine-grained attributes, spatial arrangements, and dynamic 
interactions---has emerged as a cornerstone for diverse downstream applications. These 
range from semantic image retrieval~\cite{radford2021learning,chun2021probabilistic} and assistive 
technologies~\cite{gurari2018vizwiz} to high-quality data synthesis for text-to-image 
generative models~\cite{betker2023improvingdalle3}.
With the rapid advancement of Vision-Language Models (VLMs)~\cite{google2026gemini31,achiam2023gpt4,openai2026gpt55,anthropic2026claudeopus48,anthropic2026claudesonnet5,bai2025qwen3vl,alibaba2026qwen35,alibaba2026qwen37plus,team2026kimik25}, generating long, detailed captions has become routine; yet \emph{evaluating} whether these captions are factually correct and sufficiently comprehensive remains an open challenge.

Existing approaches to caption evaluation fall into three categories, each with significant limitations.
\textbf{First}, n-gram-based metrics such as CIDEr~\cite{vedantam2015cider}, SPICE~\cite{anderson2016spice}, and BERTScore~\cite{DBLP:conf/iclr/ZhangKWWA20} measure surface-level text overlap with reference captions, but lexical similarity does not imply factual correctness---a caption can share many n-grams with a reference while attributing wrong colors to wrong objects, hallucinating nonexistent entities, or incorrectly negating visual facts.
These metrics are particularly ill-suited for detailed captions, where the space of valid descriptions is vast and n-gram overlap is a poor proxy for semantic accuracy.
\textbf{Second}, LLM-score methods~\cite{lu2025benchmarkingcomprecap,liu2026capability,cheng2025caparena} employ models to score captions based on some predefined dimensions, but these scores are sensitive to prompt wording, difficult to reproduce, costly, and---critically---lack ground truth for calibration, as the reference captions themselves are often generated by GPT-style models.
Moreover, LLM judges exhibit strong model-specific biases (\eg, favoring longer or more fluent text regardless of factual accuracy), making cross-model comparison unreliable.
\textbf{Third}, QA-based evaluation~\cite{lu2025benchmarkingcomprecap,yang2026captionqa} converts visual understanding assessment into question answering, offering better grounding than text similarity.
However, existing QA-based benchmarks leave several complementary opportunities for denser and more localized evaluation: their domain taxonomies are relatively compact, their probe density is lower than that of \method, and individual questions are generally posed at the image level rather than explicitly aligned with segmented semantic regions.
Even when dozens of questions are available for an image, image-level QA may not reveal which foreground or background regions are systematically under-described.
A caption that correctly answers a limited set of image-level questions may still contain extensive hallucinations or omissions on the hundreds of facts that were never asked about.

Our key insight stems from a \textbf{training--evaluation symmetry}: on the training side, dense caption supervision---where every region of an image is paired with detailed textual descriptions---has proven essential for learning strong vision--language alignment~\cite{chen2015microsoftmscoco,DBLP:conf/nips/0016WLD0ZCDB00024sharegpt4video,zheng2024dreamlip,li2025denseworld,johnson2016densecap,wu2024grit,lian2025describeanything,you2024ferret,gao2022caponimage}.
If a model trained with dense supervision should ideally produce captions that fully align with the image, then evaluation should operate at the same granularity.
This motivates us to construct QA pairs directly from image content at the region level, enabling denser and more localized factual checks than image-level QA alone.

To this end, we introduce \method, a \textbf{full-scene dense QA benchmark} for evaluating detailed image captions, where ``full-scene'' denotes a design goal of covering both foreground and background regions rather than a guarantee of exhaustive instance coverage.
To achieve broad scene coverage, \method adopts a region-centric decomposition approach, partitioning each image into coarse \textbf{semantic regions}---encompassing localized foreground objects, parts, and background elements.
By generating detailed QA pairs for every retained region across 10 predefined semantic categories, \method constructs a dense factual checklist over the extracted regions to evaluate whether captions support the probed visual information.
\method is constructed through a principled pipeline: we first select images spanning \textbf{37 L1 domains} and \textbf{219 L2 sub-domains} from LVIS~\cite{gupta2019lvis}, Places365~\cite{zhou2017places}, and OpenImagesV7~\cite{kuznetsova2020openimagesv7} for broad taxonomy coverage; then segment each image into semantic regions using YOLOv26-seg~\cite{jocher2026ultralyticsyolo26} and SAM3~\cite{carion2025sam3}; generate structured metadata for every region via Gemini-3.1-Pro~\cite{google2026gemini31}; and finally produce dense multiple-choice questions with Gemini-3.1-Pro and GPT-5.5~\cite{openai2026gpt55} that probe the retained regions and their metadata across the predefined QA categories.
The resulting benchmark contains \textbf{346 images}, \textbf{1,868 extracted regions}, and \textbf{25,650 QA pairs}---averaging \textbf{74 QA pairs per image} and \textbf{13.7 QA pairs per region}, compared with 50.3 QA pairs per image in CaptionQA, while additionally providing explicit region-level anchoring.
The evaluation proceeds in three stages: (1)~caption generation, (2)~caption-based QA answering with an Uncertain option, and (3)~multi-dimensional metric computation.
Our evaluation protocol is cost-effective and reduces open-ended LLM-as-scorer bias by grounding assessment in verifiable multiple-choice answers, while remaining dependent on the language reader used at evaluation time.
Furthermore, the Uncertain option and \emph{Effective Accuracy} provide a judge-dependent proxy for distinguishing unanswered probes from incorrectly resolved ones, while \emph{density-based metrics} penalize verbose yet uninformative captions.

Our main contributions are summarized as follows:
\begin{enumerate}[leftmargin=*,itemsep=2pt,topsep=3pt]
    \item \textbf{Decomposition Paradigm \& Benchmark:} We propose a novel evaluation paradigm for detailed captions based on region-centric scene decomposition, realized via \textbf{\method}. By deploying region-level QA as fine-grained factual probes across 10 semantic categories, \method increases question density (avg.~74 QA/image) while explicitly grounding each probe in a retained semantic region to rigorously cross-examine detailed captions over the retained regions.
    \item \textbf{Taxonomy-Guided Selection:} We design a two-tier taxonomy (37 L1 domains and 219 L2 sub-domains) to curate a taxonomy-diverse image suite spanning both common and long-tail tags, without claiming statistical representativeness of real-world visual distributions.
    \item \textbf{Multi-Dimensional Metrics \& Protocol:} We introduce a multi-dimensional evaluation protocol with competency metrics (Overall/Effective Accuracy, Uncertain Ratio, Coverage) and efficiency metrics (Global Density, Density CV).
    \item \textbf{Extensive Evaluation:} We benchmark 13 state-of-the-art VLMs on CapProbe and analyze competency–efficiency trade-offs that remain invisible under less granular or overlap-based evaluation.
\end{enumerate}

%% file: sec/2_related_work.tex
\section{Related Work}\label{sec:related}

\subsection{Image Captioning and Benchmarks}\label{sec:rw_captioning}

Image captioning has been a core vision-language task for over a decade.
Early benchmarks such as MSCOCO Captions~\cite{chen2015microsoftmscoco} and Flickr30k~\cite{young-etal-2014-imageflickr30} focus on short descriptions, which are insufficient for evaluating the long, detailed captions produced by modern VLMs.
NoCaps~\cite{agrawal2019nocaps} extends coverage to novel object categories but retains the short-caption format.
In a more recent attempt, DetailCaps~\cite{dong2024benchmarkingcapture} assesses detailed captions by extracting and comparing key semantic terms from candidate and reference texts, though this approach remains limited to shallow keyword-level matching.
CompreCap~\cite{lu2025benchmarkingcomprecap} provides scene-graph-annotated images with attribute and relation labels, enabling structured evaluation, but its QA component remains sparse and its domain coverage is limited.
CaptionQA~\cite{yang2026captionqa} provides a dense, utility-oriented benchmark with 50.3 questions per image across four domains, but its questions are posed at the image level rather than explicitly anchored to segmented semantic regions.
CAPability~\cite{liu2026capability} adopts a capability-oriented evaluation framework but does not perform dense, region-level probing of caption quality.
\method complements these benchmarks by increasing probe density to 74 questions per image, explicitly grounding each question in a retained semantic region, and broadening taxonomy coverage to 37 L1 and 219 L2 domains.
\subsection{Caption Evaluation Metrics}\label{sec:rw_metrics}

Caption evaluation metrics can be broadly categorized into three families.

\noindent\textbf{N-gram-based metrics.}
BLEU~\cite{papineni-etal-2002-bleu}, ROUGE~\cite{lin-2004-rouge}, METEOR~\cite{banerjee-lavie-2005-meteor}, CIDEr~\cite{vedantam2015cider}, and SPICE~\cite{anderson2016spice} measure text similarity between generated and reference captions.
While efficient and reference-based, they conflate lexical overlap with semantic correctness and are particularly ill-suited for detailed captions, where multiple valid descriptions may share few n-grams.
SPICE partially addresses this by parsing scene graphs, but isolated matching of objects, attributes, and relations disrupts their inherent binding.
Moreover, these metrics cannot evaluate negation, relational accuracy, or attribute--region correspondence---all critical for detailed captions.

\noindent\textbf{Embedding-based metrics.}
CLIPScore~\cite{hessel2021clipscore} leverages vision-language embeddings to measure image--caption alignment without references.
While more robust to paraphrasing, they provide only a single scalar score and cannot diagnose \emph{which} aspects of a caption are correct or incorrect.

\noindent\textbf{LLM-as-Judge.}
Recent works employ LLMs such as GPT~\cite{hurst2024gpt4o} or Llama~\cite{touvron2023llama,grattafiori2024llama3} to score captions.
However, these approaches suffer from inconsistent scoring across runs, high API costs, and the absence of ground truth for calibration---the reference captions used for scoring are often themselves generated by GPT-style models, creating a circular evaluation.
More importantly, LLM judges exhibit strong model-specific biases (\eg, systematically favoring longer or more fluent text), making cross-model comparison unreliable and hard to reproduce.
Recent works~\cite{yang2026captionqa} start to avoid scoring directly and instead use the LLM as a \emph{reader} (answering MCQs from captions) rather than a \emph{scorer}, grounding evaluation in verifiable question--answer pairs with objective correctness criteria.
This paradigm is cost-effective, reduces open-ended LLM-as-scorer bias relative to unconstrained scalar scoring, and supports reproducible comparative ranking under a fixed reader---while absolute scores remain reader-dependent.

\subsection{QA-based Caption and VLM Evaluation}\label{sec:rw_qa}

QA-based evaluation converts the assessment of visual understanding into question answering.
POPE~\cite{li2023evaluatingpope} probes object hallucination through yes/no questions on object presence, but covers only a single dimension.
Q-Bench~\cite{wu2024qbench} constructs one human-asked question per image for image quality assessment, yet its QA density remains low and questions are not anchored to specific visual regions.
FGHE~\cite{wang2024mitigatingfghe} extends POPE with fine-grained hallucination evaluation but retains the yes/no format, limiting diagnostic granularity.
CompreCap~\cite{lu2025benchmarkingcomprecap} designs VQA tasks for tiny objects with A/B/C choices, but its QA pairs cover only a fraction of the visual content.
More recently, CaptionQA~\cite{yang2026captionqa} introduces dense image-level QA for measuring caption utility, while CapRL~\cite{xing2025caprl} uses QA-based signals to improve dense caption generation.
Neither explicitly associates every evaluation probe with a segmented semantic region.
In contrast, \method generates dense QA pairs, each anchored to a specific semantic region, enabling fine-grained probing of caption quality over the extracted region set.
This explicit region alignment establishes a localized vision--language correspondence beyond image-level QA alone.

\subsection{Dense and Region-level Supervision}\label{sec:rw_dense}

The vision-language community has long recognized the importance of dense, region-level signals for \emph{training}.
DenseCap~\cite{johnson2016densecap} introduces dense captioning, where every region is described with a short phrase.
GRIT~\cite{wu2024grit} and other region-level captioning methods~\cite{lian2025describeanything,you2024ferret} demonstrate that region-grounded supervision significantly improves captioning quality.
In the pretraining era, large-scale image--caption alignment datasets~\cite{chen2015microsoftmscoco,DBLP:conf/nips/0016WLD0ZCDB00024sharegpt4video,zheng2024dreamlip,li2025denseworld} pair images with long, detailed descriptions to teach VLMs fine-grained vision--language correspondence.
Yet evaluation has not kept pace: while models are trained with dense signals, they are evaluated with sparse metrics.
\method closes this \emph{training--evaluation gap} by providing dense, region-aligned QA at evaluation time, ensuring that the granularity of assessment matches the granularity of the captions being produced.

%% file: sec/3_method.tex
\section{The CapProbe Benchmark}\label{sec:method}

\begin{figure*}[t]
    \centering
    \includegraphics[width=\textwidth]{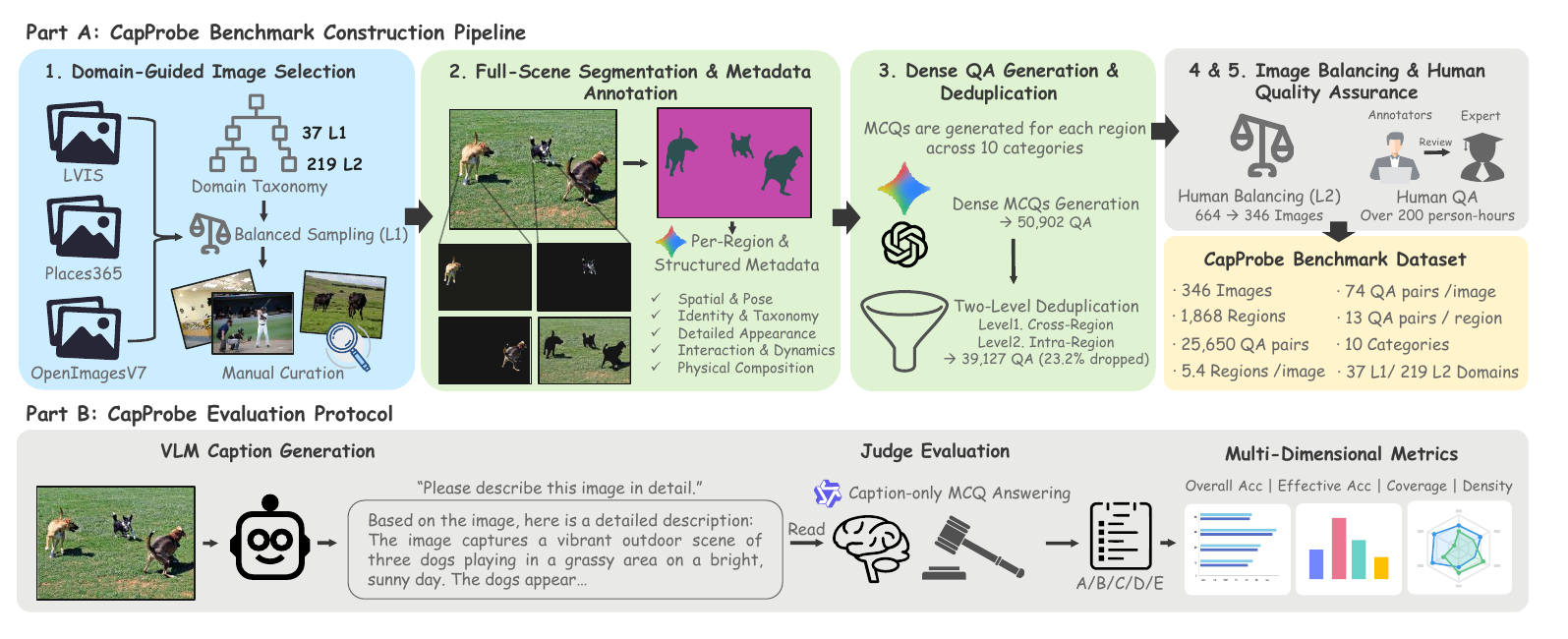}
    \caption{Construction and evaluation pipeline of \method.
    We select images under a hierarchical domain taxonomy, segment semantic regions, annotate structured metadata, generate and deduplicate dense QA pairs, balance domain coverage, and apply human quality assurance; captions are then evaluated via caption-based QA answering.}
    \label{fig:pipeline}
\end{figure*}

A high-quality detailed caption should not only mention the objects present in an image but also faithfully describe their attributes, spatial relationships, and interactions, covering \emph{all} visual content, not just salient foreground elements.
However, existing caption evaluation benchmarks either rely on shallow text similarity or use QA probes without explicitly localizing each question to a semantic image region, leaving most fine-grained facts unchecked.
To bridge this gap, we propose \method, a \textbf{full-scene dense question-answering} benchmark that evaluates detailed image captions by probing how many fine-grained visual facts a caption can support over a set of extracted semantic regions.
As illustrated in~\cref{fig:pipeline}, the construction of \method follows a five-stage pipeline: (1)~hierarchical domain-guided image selection (\cref{sec:image_selection}), (2)~semantic region segmentation and metadata annotation (\cref{sec:segmentation},~\cref{sec:metadata}), (3)~dense QA generation and deduplication (\cref{sec:qa_generation},~\cref{sec:dedup}), (4)~balanced image selection (\cref{sec:image_balancing}), and (5)~human quality assurance (\cref{sec:qa_qc}).
We present the dataset statistics in~\cref{sec:dataset_stats} and the evaluation protocol in~\cref{sec:eval_protocol}.

% ======================================================================
\subsection{Hierarchical Domain-Guided Image Selection}\label{sec:image_selection}

To ensure broad taxonomy coverage rather than statistical representativeness of the open-world image distribution, we construct a two-level domain taxonomy and select images that span it.

\begin{figure*}[t]
    \centering
    \includegraphics[width=\textwidth]{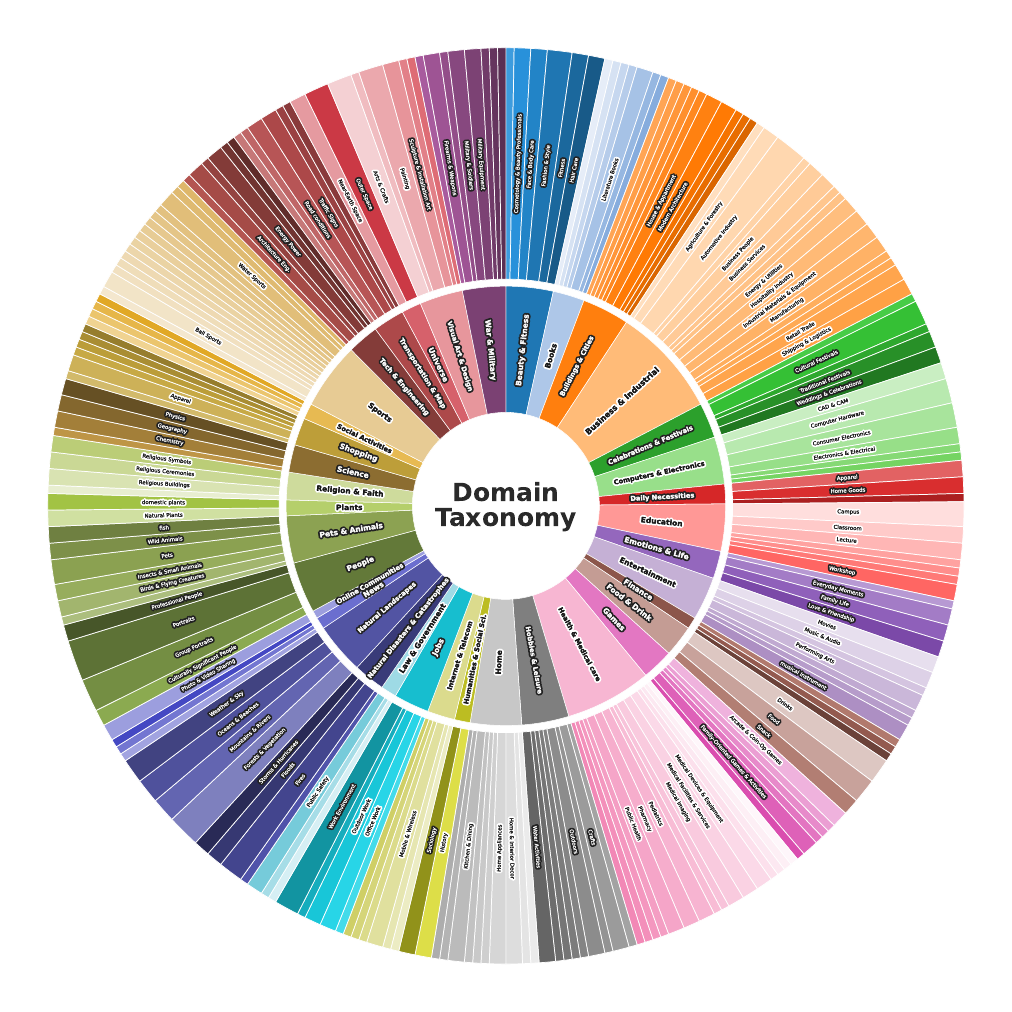}
    \caption{Hierarchical domain coverage of \method. 
    The inner ring shows 37 L1 domains and the outer ring shows 219 L2 sub-domains; sector area at each level is proportional to the number of benchmark images.
    The complete hierarchical tag system is provided in~\cref{tab:domain_taxonomy}.}
    \label{fig:domain_coverage}
\end{figure*}

\noindent\textbf{Domain taxonomy.}
We define a hierarchical tag system comprising \textbf{37 L1 domains} and \textbf{219 L2 sub-domains}.
The L1 categories span a wide spectrum of visual scenarios---from \emph{Natural Landscapes}, \emph{Sports}, and \emph{Food \& Drink} to \emph{War \& Military}, \emph{Religion \& Faith}, and \emph{Science}---ensuring that \method is not biased toward any single visual domain.
Each L1 domain is further subdivided into semantically coherent L2 sub-domains (\eg, \emph{Sports} $\rightarrow$ \emph{Ball Sports}, \emph{Water Sports}, \emph{Combat Sports}, \etc).
\cref{fig:domain_coverage} visualizes the full taxonomy. 

\noindent\textbf{Image sourcing and filtering.}
We draw candidate images from three large-scale open datasets: Places365~\cite{zhou2017places}, LVIS~\cite{gupta2019lvis}, and OpenImagesV7~\cite{kuznetsova2020openimagesv7}.
The selection proceeds in four steps:
\begin{enumerate}[leftmargin=*,itemsep=2pt,topsep=3pt]
    \item \textbf{Zero-level coarse filtering.} We remove images with low tag confidence, low resolution, or poor visual quality from each source dataset.
    \item \textbf{L1 domain matching.} Using Qwen3-VL-32B-Instruct~\cite{bai2025qwen3vl}, we match each candidate image against the 37 L1 domain labels and retain images with high matching confidence.
    \item \textbf{L2 sub-domain matching.} For the L1-matched images, we further use Qwen3-VL-32B-Instruct~\cite{bai2025qwen3vl} to assign L2 sub-domain labels, again retaining only high-confidence matches.
    \item \textbf{Balanced sampling.} We sample images per L1 domain with a target of balanced coverage, prioritizing those with the highest confidence scores.
\end{enumerate}

\noindent\textbf{Manual curation.}
After automated selection, human annotators review each image against three criteria:
(1)~\emph{Correctness}---whether the image genuinely matches its assigned L1/L2 labels;
(2)~\emph{Image quality}---whether the image is sufficiently sharp, well-exposed, and free of artifacts;
(3)~\emph{Content reasonableness}---whether the main subject is well-positioned and the scene composition is natural.
Images failing any criterion are removed.
This process yields 664 images spanning all 37 L1 domains.

% ======================================================================
\subsection{Full-Scene Segmentation}\label{sec:segmentation}

Rather than treating an image as a monolithic entity, we decompose it into semantic regions---encompassing both localized foreground objects and background elements---so that evaluation can probe fine-grained facts beyond the salient foreground alone.
This is a deliberate design choice: a detailed caption should describe not only salient foreground objects but also background context (\eg, sky, walls, ground), and our evaluation should probe these as well.
Importantly, our regions are \emph{coarse semantic parts} rather than an exhaustive inventory of object instances: a region may be a whole object, a salient part, or a background/stuff area with approximate spatial support.
We do not require ultra-fine instance splits or perfectly correct segmentation labels; the mask mainly localizes where to look, while subsequent metadata annotation supplies and corrects the factual description of that area (\cref{sec:metadata}).
``Full-scene'' therefore names a design goal---including foreground and background cues in the region set---not a verified claim that every salient entity is segmented.
To this end, \method adopts a region-centric decomposition that combines YOLOv26-seg~\cite{jocher2026ultralyticsyolo26} with SAM3~\cite{carion2025sam3}.
YOLOv26-seg produces masks for discrete objects (``things''), while SAM3 complements coverage with additional segments for background and amorphous regions (``stuff''); together they form the region set used for subsequent annotation.
We choose this approach for two reasons:
(1)~Semantic regions are natural units of visual description---a detailed caption describes \emph{which} regions exist, \emph{what} they look like, and \emph{how} they relate to each other;
(2)~Region-level segmentation enables us to associate each QA pair with a specific visual area, grounding the evaluation in localized visual evidence.
On average, each image contains 5.4 extracted regions; density in \method comes primarily from many QA probes per region rather than from maximizing the region count.

% ======================================================================
\subsection{Structured Metadata Annotation}\label{sec:metadata}

For every extracted region, we use \textbf{Gemini-3.1-Pro}~\cite{google2026gemini31} to generate a structured metadata description conditioned on both the original image and the region's segmentation mask, serving as a grounded factual anchor for subsequent question generation.
The prompt (see~\cref{sec:appendix_meta_prompt}) instructs the model to identify whether the segment is a discrete object (``thing'') or a background/environmental region (``stuff''), and to tailor the annotation granularity accordingly.
Crucially, rather than directly prompting the model to synthesize QA pairs straight from raw visual segments, we decouple the workflow into a two-stage paradigm: \emph{Image $\rightarrow$ Metadata $\rightarrow$ QA}. 
Pre-determining regional characteristics into a structured format provides exceptional semantic stability and clear contextual constraints. 
By conditioning the final QA generation on this verified metadata alongside the visual inputs, we significantly eliminate open-ended hallucination and enhance question accuracy, yielding substantially higher stability than naive direct-generation alternatives.
The metadata schema (see~\cref{fig:meta_prompt}) is designed to capture the full spectrum of visual information that a detailed caption may convey, organized into five modules:

\begin{itemize}[leftmargin=*,itemsep=2pt,topsep=3pt]
    \item \textbf{Spatial \& Pose:} bounding box, global position (\eg, ``left foreground''), visibility state (occlusion, visible range, out-of-frame parts), view angle, orientation, and physical scale.
    \item \textbf{Identity \& Taxonomy:} broad category, specific type, and primary function.
    \item \textbf{Detailed Appearance:} key identifying feature, overall color palette, component-level features (part name, description, visibility, quantity), text/markings (transcribed content, type, placement), surface texture, and specific details.
    \item \textbf{Interaction \& Dynamics:} interaction targets (with target name, count, interaction type, and detail), motion state, current action, and condition.
    \item \textbf{Physical Composition:} materials and structural integrity.
\end{itemize}

This schema records a broad set of attributes that a detailed caption may convey for the region---from color and texture to interactions with other regions---providing a structured foundation for dense QA generation.

% ======================================================================
\subsection{Dense QA Generation}\label{sec:qa_generation}

Given the structured metadata for each region, we employ \textbf{Gemini-3.1-Pro}~\cite{google2026gemini31} and \textbf{GPT-5.5}~\cite{openai2026gpt55} to generate multiple-choice questions (MCQs) with four answer options (A--D), using the same prompt for both generators.
The \textbf{Uncertain} option (E) is introduced later at evaluation time (\cref{sec:eval_protocol}), so that a language judge can decline to answer when the caption lacks sufficient evidence.
The prompt (see~\cref{sec:appendix_qa_prompt}) enforces three key design principles: (1)~\emph{zero ambiguity}---every question must include a unique identifying description based on the region's position and features; (2)~\emph{natural language}---no technical jargon such as ``masked region''; (3)~\emph{exhaustive coverage}---questions must span all QA categories.
\begin{figure}[t]
    \centering
    \includegraphics[width=\columnwidth]{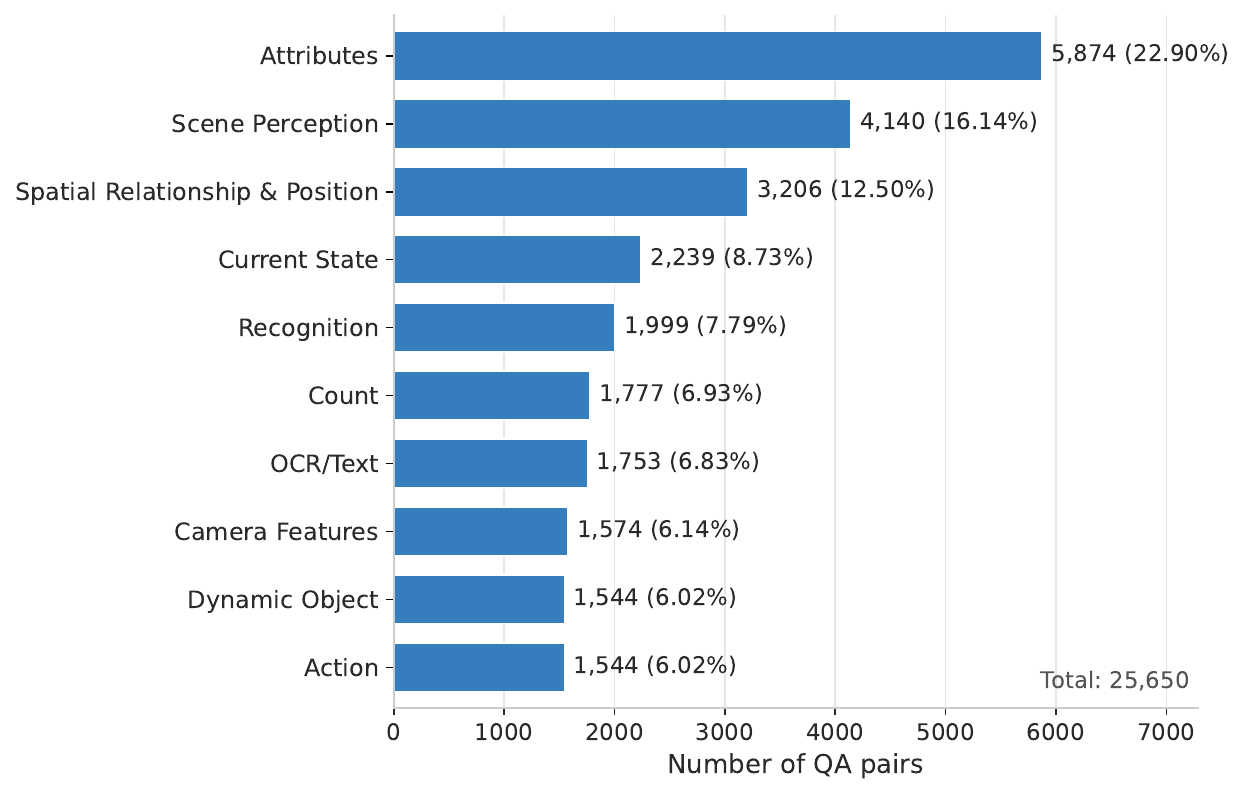}
    \caption{Distribution of QA pairs across 10 categories.}
    \label{fig:qa_category}
\end{figure}

\begin{itemize}[leftmargin=*,itemsep=2pt,topsep=3pt]
    \item \textbf{Attributes}: color, shape, texture, material, \etc
    \item \textbf{Scene Perception}: overall scene type, lighting, atmosphere
    \item \textbf{Spatial Relationship \& Position}: relative position, occlusion, depth ordering
    \item \textbf{Current State}: condition, motion state, ongoing action
    \item \textbf{Recognition}: region category, specific type identification
    \item \textbf{Count}: quantity of regions or region parts
    \item \textbf{OCR/Text}: text content, signage, labels
    \item \textbf{Camera Features}: viewpoint, angle, framing
    \item \textbf{Dynamic Object}: moving elements, interaction dynamics
    \item \textbf{Action}: specific actions performed by agents
\end{itemize}

% ======================================================================
\subsection{Deduplication}\label{sec:dedup}

Automated QA generation can produce semantically redundant questions, both within a single region and across different regions of the same image.
We therefore apply a two-level hierarchical deduplication strategy based on semantic embedding similarity.

\noindent\textbf{Embedding model.}
We encode each QA pair by concatenating the question text with all answer options, and compute its embedding using Qwen3-Embedding-8B~\cite{zhang2025qwen3emb}, similar to~\cite{yang2026captionqa}.
All embeddings are L2-normalized, and pairwise cosine similarity is computed for every QA pair within the same image.

\noindent\textbf{Two-level deduplication.}
We adopt a hierarchical strategy with two thresholds (we sampled a subset to determine the thresholds with human verification):
\begin{enumerate}[leftmargin=*,itemsep=2pt,topsep=3pt]
    \item \textbf{Level 1---Cross-region deduplication} ($\tau_{\text{global}} = 0.90$): We compare all QA pairs across \emph{all regions} within the same image. If a pair exceeds the global threshold, the later one is removed. This eliminates near-duplicate questions that arise when multiple regions share similar visual content (\eg, two objects of the same category with similar attributes).
    \item \textbf{Level 2---Intra-region deduplication} ($\tau_{\text{local}} = 0.85$): Among the surviving QA pairs, we further compare those belonging to the \emph{same region} under a more lenient threshold. This catches subtler redundancies within a single region's QA set, where questions may probe the same fact from slightly different angles.
\end{enumerate}
When a duplicate is detected, we retain the first occurrence and remove the later one, since all retained QA pairs subsequently undergo human quality assurance (\cref{sec:qa_qc}).
If two highly similar questions have different ground-truth answers, both are kept to preserve distinct factual content.

\noindent\textbf{Statistics.}
Starting from 50,902 raw QA pairs, Level~1 removes 3,268 cross-region duplicates and Level~2 removes 8,507 intra-region duplicates, yielding \textbf{39,127 QA pairs} after deduplication.
This 23.2\% reduction confirms that automated generation introduces substantial redundancy, validating the necessity of deduplication.

% ======================================================================
\subsection{Image Balancing}\label{sec:image_balancing}

After deduplication, the image pool still contains an uneven distribution across L2 sub-domains---some sub-domains have many more candidate images than others.
To balance taxonomy coverage, we manually cap the number of images per L2 sub-domain at \textbf{5}, selecting 1--5 images per sub-domain based on image quality and label fit.
Most L2 sub-domains retain only 1 or 2 images, so the suite is taxonomy-diverse but does not support stable per-L2 performance estimates.
This balancing step reduces the dataset from 664 images and 39,127 QA pairs to \textbf{346 images}, \textbf{1,868 extracted regions}, and \textbf{26,062 QA pairs}, ensuring that no single L1 domain dominates the evaluation by image count.

% ======================================================================
\subsection{Human Quality Assurance}\label{sec:qa_qc}

The final stage of our pipeline involves human verification of the automatically generated QA pairs.
Three trained annotators spent over a total of \textbf{200 person-hours} reviewing all 26,062 QA pairs, with the workload split approximately evenly across annotators.
For every pair, the assigned annotator inspects the original image together with the corresponding region mask and assigns one of three decisions: \emph{accept}, \emph{edit}, or \emph{delete}.
Each decision is made along the following dimensions:
(1)~\emph{Factual correctness}---whether the ground-truth answer accurately reflects the visual content of the corresponding region;
(2)~\emph{Question clarity}---whether the question is unambiguous and answerable from the image alone;
(3)~\emph{Option quality}---whether the distractor options are plausible yet clearly distinguishable from the correct answer;
(4)~\emph{Redundancy}---whether the question probes a distinct fact not already covered by another QA pair in the same image.
All edited pairs, together with a random 5\% sample of accepted pairs, are then reviewed by an expert adjudicator, whose judgment is final.
After this process, the final benchmark contains \textbf{25,650 QA pairs}---a 1.6\% reduction from 26,062 corresponding to pairs marked for deletion.

% ======================================================================
\subsection{Dataset Statistics}\label{sec:dataset_stats}

\cref{fig:pipeline} summarizes the key statistics of \method.
Compared to existing caption evaluation benchmarks, \method is distinguished by its \textbf{density}---74 QA pairs per image---and its \textbf{domain breadth}---37 L1 domains covering natural landscapes, urban scenes, human activities, scientific imagery, and more.
We provide a detailed comparison with prior benchmarks in~\cref{tab:benchmark_comparison}.

On average, each region receives \textbf{13.7 QA pairs}, and each image accumulates \textbf{74 QA pairs}---a higher probe density than prior QA-based caption benchmarks such as CaptionQA (50.3 QA pairs per image), together with explicit region-level anchoring.
Holding probe quality constant, evaluating a caption against 74 questions checks a broader set of visual facts than evaluating it against a smaller image-level question set.
By constructing QA pairs for every retained region, \method builds a \textbf{dense factual checklist} that evaluates whether a caption supports the visual information probed over those regions.

\begin{table*}[t]
    \centering
    \caption{Comparison of \method with existing image caption evaluation benchmarks.
    Human QC: \emph{Ref only} = reference captions only; \emph{Caption/SG/Dim/QA annot.} = single-stage human annotation; Multi-stage = review at both image selection and QA verification.
    Probes/Img: factual probes per image. Region: whether probes are anchored to segmented regions.}
    \label{tab:benchmark_comparison}
    \scriptsize
    \begin{threeparttable}
    \begin{tabular}{@{}lcccccccc@{}}
        \toprule
        Benchmark & \#Img & Domain Coverage & Human QC & Probes/Img & Region & Omit/Hallu. & Metrics \\
        \midrule
        MSCOCO$_{\text{val}}$~\cite{chen2015microsoftmscoco}
            & 5{,}000 & COCO & Ref only & 5 refs & $\times$ & $\times$ & Overlap \\
        NoCaps$_{\text{val}}$~\cite{agrawal2019nocaps}
            & 4{,}500 & COCO+novel & Ref only & 11 refs & $\times$ & $\times$ & Overlap \\
        POPE~\cite{li2023evaluatingpope}
            & 500 & COCO & $\times$ & 6 & $\times$ & $\times$ & Binary acc \\
        DetailCaps~\cite{dong2024benchmarkingcapture}
            & 100 & Custom & Caption & 1 ref & $\times$ & $\times$ & Scalar score \\
        CompreCap~\cite{lu2025benchmarkingcomprecap}
            & 560 & COCO/Flickr & SG annot. & Sparse & partial & $\times$ & SG match; sparse MCQ \\
        CAPability~\cite{liu2026capability}
            & ${\sim}$11K & 6 views & Dim annot. & Sparse & $\times$ & partial & Per-dim score \\
        CaptionQA~\cite{yang2026captionqa}
            & 657 & 4 domains & QA annot. & 50 & $\times$ & $\checkmark$ & MCQ acc \\
        \midrule
        \textbf{\method} (Ours)
            & 346 & \textbf{37 L1, 219 L2} & \textbf{Multi-stage} & \textbf{74} & $\checkmark$ & $\checkmark$ & \textbf{MCQ acc + abst./err.\ proxy + density-based} \\
        \bottomrule
    \end{tabular}
    \end{threeparttable}
\end{table*}

% ======================================================================
\subsection{Evaluation Protocol}\label{sec:eval_protocol}

Given a generated caption for an image, \method evaluates its quality in three stages:

\noindent\textbf{Stage 1: Caption generation.}
Each caption model under evaluation generates a detailed caption for every image in \method using a unified prompt:
\begin{quote}
    \small\textit{``Please describe this image in detail.''}
\end{quote}

\noindent\textbf{Stage 2: Caption-based QA answering.}
A language judge model reads \emph{only} the generated caption (without access to the image) and answers all QA pairs associated with that image.
Each question offers five options: A, B, C, D, and E (Not mentioned / Cannot be determined).
The caption is wrapped in \texttt{<CAPTION>} tags, and the judge is instructed to use \textbf{only} the information within those tags, selecting the uncertain option E when the caption lacks sufficient information (full prompt in~\cref{sec:appendix_judge_prompt}).

\noindent\textbf{Stage 3: Multi-dimensional metric computation.}
Based on the judge's answers, we compute competency metrics (Overall Accuracy, Effective Accuracy, Uncertain Ratio, Coverage) and efficiency metrics (Tokens/Image, Mean Density, Global Density, Density CV).
The full definitions and design rationale of each metric are presented in~\cref{sec:metrics}.

\noindent\textbf{Design rationale for the Uncertain option.}
Without option~E, a judge forced to choose among A--D may guess when the caption lacks the relevant information, conflating unanswered probes with incorrectly resolved ones.
The Uncertain option enables \emph{Effective Accuracy}, measured only on questions the judge commits to answering, providing a judge-dependent proxy for this distinction rather than a causal measure of caption hallucination versus omission.

%% file: sec/4_metrics.tex
\section{Evaluation Metrics}\label{sec:metrics}

We evaluate captions by measuring how well a language judge can answer \textbf{dense, region-aligned multiple-choice questions} using \textbf{only} the generated caption.
Beyond binary accuracy, we introduce an \textbf{Uncertain} option that yields a judge-dependent proxy separating unanswered probes from incorrectly resolved answers, and \textbf{density-based metrics} to account for caption length and information efficiency.
Our metrics are organized into two dimensions: \emph{competency} (\cref{sec:competency}) and \emph{efficiency} (\cref{sec:efficiency}).

% ======================================================================
\subsection{Competency Metrics}\label{sec:competency}

Competency metrics measure the \emph{accuracy} and \emph{attitude} of the judge when answering questions based solely on the generated caption.

\noindent\textbf{Notation.}
Let $\mathcal{I}$ denote the set of $N$ test images.
For image $i$, let $\mathcal{Q}_i$ be its associated QA set with $M_i$ questions, and let $M = \sum_i M_i$ be the total number of questions.
Let $c_i$ denote the generated caption for image $i$, and $T_i$ its token count.
For question $j$ of image $i$, the judge outputs $y_{ij} \in \{A, B, C, D, E\}$, where $E$ denotes \emph{Uncertain}; the ground-truth answer is $g_{ij} \in \{A, B, C, D\}$.

We define:
\begin{itemize}[leftmargin=*,itemsep=2pt,topsep=3pt]
    \item $C$: total number of correctly answered questions ($y_{ij} = g_{ij}$).
    \item $U$: total number of questions where the judge selects $E$.
\end{itemize}

\noindent\textbf{Overall Accuracy} measures the global correctness across all questions:
\begin{equation}\label{eq:overall}
    \text{Overall Acc} = \frac{C}{M}
\end{equation}
where selecting $E$ is counted as incorrect.

\noindent\textbf{Effective Accuracy} measures the accuracy only on questions the judge \emph{commits to answering}:
\begin{equation}\label{eq:effective}
    \text{Effective Acc} = \frac{C}{M - U} \quad (M - U > 0)
\end{equation}
This metric summarizes how often committed answers match the ground truth; it is not a direct causal measure of caption factuality, since judge errors, caption ambiguity, and lucky guesses can all affect $C$.

\noindent\textbf{Uncertain Ratio} quantifies the proportion of questions the judge declines to answer:
\begin{equation}\label{eq:uncertain}
    \text{Uncertain Ratio} = \frac{U}{M}
\end{equation}
A high Uncertain Ratio indicates frequent judge abstention, which may reflect caption omissions but can also arise from a conservative reader.

\noindent\textbf{Coverage} is the complement of Uncertain Ratio, measuring the proportion of questions receiving a substantive (non-$E$) answer:
\begin{equation}\label{eq:coverage}
    \text{Coverage} = \frac{M - U}{M} = 1 - \text{Uncertain Ratio}
\end{equation}
By definition, these quantities are not independent:
\begin{equation}\label{eq:overall_factorization}
    \text{Overall Acc} = \text{Effective Acc} \times \text{Coverage}.
\end{equation}
Overall Accuracy is therefore the product of how often the judge answers and how often those answers are correct.

\noindent\textbf{Diagnostic interpretation.}
The interplay between Overall Accuracy, Effective Accuracy, and Uncertain Ratio reveals judge-mediated response patterns under a fixed reader:
\begin{itemize}[leftmargin=*,itemsep=2pt,topsep=3pt]
    \item \emph{Effective $\approx$ Overall}: the judge rarely selects Uncertain.
    \item \emph{Effective $\gg$ Overall}: high abstention with relatively accurate committed answers (a conservative pattern under this judge).
    \item \emph{Low Overall + Low Uncertain}: the judge commits often but is frequently wrong (unsupported or incorrectly resolved information under this judge).
\end{itemize}
These patterns are useful proxies, not validated labels of caption-level hallucination versus omission.

% ======================================================================
% ======================================================================
\subsection{Efficiency Metrics}\label{sec:efficiency}

Dense QA evaluation naturally raises the question: \emph{is a longer caption necessarily better?}
A verbose caption that answers the same number of questions correctly as a concise one is less efficient.
We therefore introduce density-based metrics to measure the \emph{information efficiency} of captions.
Let $C_i$ denote the number of correctly answered questions for image $i$ (so $C=\sum_i C_i$).
Because Stage~2 answers one question at a time from the same caption, counting token cost at the QA-sample level multiplies the caption length by the number of probes: the request-token mass of image $i$ is $M_i T_i$.

\noindent\textbf{Per-image information density.}
\begin{equation}\label{eq:density}
    d_i = \frac{C_i}{M_i T_i} \times 1000
\end{equation}
Equivalently, $d_i = (C_i/M_i)\,/\,T_i \times 1000$: the image-level Overall Accuracy per thousand caption tokens.
Reported density values are in per-mille (\textperthousand).

\noindent\textbf{Tokens per Image} measures the average \emph{unique} caption length (each image counted once):
\begin{equation}\label{eq:tokens}
    \bar{T} = \frac{1}{N}\sum_{i=1}^{N} T_i
\end{equation}

\noindent\textbf{Mean Density vs.\ Global Density.}
The two density summaries are \emph{not} interchangeable: Mean Density asks ``how efficient is a typical image?'', while Global Density asks ``how efficient is the entire evaluation under request-token accounting?''.

\noindent\textbf{Mean Density} is the \emph{unweighted} arithmetic mean of $\{d_i\}$---every image contributes equally, regardless of how many QA pairs or tokens it consumes:
\begin{equation}\label{eq:mean_density}
    \bar{d} = \frac{1}{N}\sum_{i=1}^{N} d_i
\end{equation}

\noindent\textbf{Global Density} pools all correct answers and all request tokens first, then takes one ratio:
\begin{equation}\label{eq:global_density}
    d_{\text{global}} = \frac{\sum_{i} C_i}{\sum_{i} M_i T_i} \times 1000 = \frac{C}{\sum_{i} M_i T_i} \times 1000
\end{equation}
Equivalently, with weights $w_i=M_i T_i$,
\begin{equation}\label{eq:global_as_weighted}
    d_{\text{global}} = \frac{\sum_{i} w_i\, d_i}{\sum_{i} w_i}.
\end{equation}
So $d_{\text{global}}$ is the \emph{$w_i$-weighted} mean of $\{d_i\}$, whereas $\bar{d}$ is the equal-weight mean.
In particular, $N\bar{d}=\sum_i d_i$ is \emph{not} equal to $d_{\text{global}}$, and $\bar{d}=d_{\text{global}}$ only when all images share the same request-token mass $w_i$.
When long and/or densely probed images have lower $d_i$, they pull $d_{\text{global}}$ below $\bar{d}$---the pattern observed throughout~\cref{tab:main_results}.

\noindent\textbf{Density CV (Coefficient of Variation)} measures the cross-image consistency of $\{d_i\}$:
\begin{equation}\label{eq:density_cv}
    \text{CV} = \frac{\sigma(d_i)}{\bar{d}}
\end{equation}
A low CV indicates stable per-image efficiency; a high CV indicates that $d_i$ fluctuates sharply with image content.

\noindent\textbf{Diagnostic interpretation.}
\begin{itemize}[leftmargin=*,itemsep=2pt,topsep=3pt]
    \item \emph{Mean}: typical per-image information efficiency (image-centric).
    \item \emph{Global}: corpus-level efficiency under QA$\times$caption token accounting (cost-centric).
    \item \emph{Mean $\approx$ Global}: $\{d_i\}$ are similar under both equal and token-mass weighting.
    \item \emph{Mean $\gg$ Global}: high-$w_i$ images are less dense (often long captions and/or many probes with few correct answers).
    \item \emph{Global $\gg$ Mean}: high-$w_i$ images are more dense.
    \item \emph{Low / High CV}: stable vs.\ volatile per-image density across the benchmark.
\end{itemize}

% ======================================================================
\subsection{Metric Summary}\label{sec:metric_summary}

\begin{table}[t]
    \centering
    \caption{Summary of \method evaluation metrics.}
    \label{tab:metric_summary}
    \scriptsize
    \setlength{\tabcolsep}{2pt}
    \begin{tabular}{@{}ll@{\hspace{4pt}}c@{\hspace{6pt}}p{0.4\columnwidth}@{}}
        \toprule
        Dim. & Metric & Formula & Intuition \\
        \midrule
        \multirow{4}{*}{Competency}
        & Overall Acc & $C/M$ & Global correctness ($=$ Eff.\ $\times$ Cov.) \\
        & Effective Acc & $C/(M{-}U)$ & Accuracy on committed answers \\
        & Uncertain Ratio & $U/M$ & Judge abstention rate \\
        & Coverage & $(M{-}U)/M$ & Non-$E$ answering rate \\
        \midrule
        \multirow{4}{*}{Efficiency}
        & Tokens/Img & $\frac{1}{N}\sum T_i$ & Avg.\ caption length \\
        & Mean Density & $\frac{1}{N}\sum d_i$ & Typical per-image efficiency \\
        & Global Density & $\frac{\sum_i w_i d_i}{\sum_i w_i}$ & Corpus efficiency ($w_i{=}M_iT_i$) \\
        & Density CV & $\sigma(d_i)/\bar{d}$ & Cross-image consistency \\
        \bottomrule
    \end{tabular}
\end{table}

\cref{tab:metric_summary} summarizes all metrics.
In the main results table (\cref{tab:main_results}), we report: Overall Acc~$\uparrow$, Effective Acc~$\uparrow$, Uncertain Ratio~$\downarrow$, Coverage~$\uparrow$, Tokens/Image, Mean Density~$\uparrow$, Global Density~$\uparrow$, and Density CV~$\downarrow$.
Together, these metrics provide a comprehensive diagnostic: competency metrics capture whether a caption truly ``holds up'' under dense probing, while efficiency metrics enable fair comparison across models with different caption length strategies.

%% file: sec/5_exp.tex
\section{Experiments}\label{sec:exp}

In this section, we benchmark a diverse set of VLMs on \method and analyze the results from multiple perspectives.
We first describe the experimental setup (\cref{sec:setup}), then present the main results (\cref{sec:main_results}), and provide in-depth analyses (\cref{sec:analysis}).

% ======================================================================
\subsection{Experimental Setup}\label{sec:setup}

\noindent\textbf{Caption models.}
We evaluate 13 representative VLMs spanning proprietary and open-weight families, including Gemini-3.1-Pro~\cite{google2026gemini31}, Qwen3.7-Plus~\cite{alibaba2026qwen37plus}, Kimi-K2.5~\cite{team2026kimik25}, Claude-Opus-4.8 and Claude-Sonnet-5~\cite{anthropic2026claudeopus48,anthropic2026claudesonnet5}, GPT-5.5 and GPT-4o~\cite{openai2026gpt55,hurst2024gpt4o}, Qwen3.5 (397B/122B/27B)~\cite{alibaba2026qwen35}, and Qwen3-VL (235B/32B/8B)~\cite{bai2025qwen3vl}.
All models generate captions in a zero-shot manner using the unified prompt.

\noindent\textbf{Judge model.}
We employ \textbf{Qwen3-32B}~\cite{yang2025qwen3} as the default language judge for Stage~2 (caption-based QA answering).
The judge receives only the generated caption and the question, and selects one of five options (A/B/C/D/E).
Limited by compute resource, we use 8-bit quantization to reduce memory usage and speedup the inference. 

\noindent\textbf{Tokenizer.}
Token counts $T_i$ are computed using o200k\_base~\cite{openai2024tiktoken} to ensure consistent cross-model comparison of density metrics.

% ======================================================================
\subsection{Main Results}\label{sec:main_results}

\begin{table*}[t]
    \centering
    \caption{Main results on \method.
    Models are grouped by license (closed-source vs.\ open-source), sorted by model family and release date (earlier first).
    Top three results in each column use cell backgrounds:
    \rankmarkfirst{1st} (red, bold), \rankmarksecond{2nd} (blue, underline), and \rankmarkthird{3rd} (green, italic).
    All accuracy and ratio values are percentages; Mean/Global Density are in \textperthousand.}
    \label{tab:main_results}
    \scriptsize
    \setlength{\tabcolsep}{3.2pt}
    \begin{tabular}{@{}llccccccccc@{}}
        \toprule
        Type & Model & Year & Overall $\uparrow$ & Effective $\uparrow$ & Uncertain $\downarrow$ & Coverage $\uparrow$ & Tokens/Img & Mean Dens.\ $\uparrow$ & Global Dens.\ $\uparrow$ & Density CV $\downarrow$ \\
        \midrule
        \multirow{6}{*}{\shortstack[l]{Closed-\\source}}
        & GPT-4o~\cite{hurst2024gpt4o}                      & 2024.05 & 34.41 & \rankfirst{94.46} & 63.57 & 36.43 & 173 & \rankfirst{2.55} & \ranksecond{1.91} & 0.569 \\
        & GPT-5.5~\cite{openai2026gpt55}                    & 2026.04 & 47.47 & \ranksecond{93.93} & 49.46 & 50.54 & 222 & \ranksecond{2.53} & \rankfirst{2.08} & 0.422 \\
        & Gemini-3.1-Pro~\cite{google2026gemini31}          & 2026.02 & \rankfirst{68.72} & 93.27 & \rankfirst{26.32} & \rankfirst{73.68} & 561 & 1.34 & 1.19 & \rankfirst{0.272} \\
        & Claude-Opus-4.8~\cite{anthropic2026claudeopus48}  & 2026.05 & 54.42 & \rankthird{93.38} & 41.72 & 58.28 & 349 & 1.75 & 1.54 & 0.320 \\
        & Claude-Sonnet-5~\cite{anthropic2026claudesonnet5} & 2026.06 & 52.97 & 93.16 & 43.14 & 56.86 & 348 & 1.73 & 1.51 & 0.351 \\
        & Qwen3.7-Plus~\cite{alibaba2026qwen37plus}         & 2026.06 & \ranksecond{63.07} & 93.23 & \ranksecond{32.35} & \ranksecond{67.65} & 433 & 1.62 & 1.43 & \ranksecond{0.307} \\
        \midrule
        \multirow{7}{*}{\shortstack[l]{Open-\\source}}
        & Qwen3-VL-8B~\cite{bai2025qwen3vl}                 & 2025.11 & 55.09 & 92.73 & 40.59 & 59.41 & 409 & 1.51 & 1.33 & 0.361 \\
        & Qwen3-VL-32B~\cite{bai2025qwen3vl}                & 2025.11 & 56.21 & 93.03 & 39.58 & 60.42 & 401 & 1.63 & 1.40 & 0.373 \\
        & Qwen3-VL-235B~\cite{bai2025qwen3vl}               & 2025.11 & 58.44 & 93.05 & 37.20 & 62.80 & 447 & 1.50 & 1.30 & 0.368 \\
        & Kimi-K2.5~\cite{team2026kimik25}                  & 2026.01 & 55.90 & 93.32 & 40.10 & 59.90 & 303 & \rankthird{2.14} & \rankthird{1.80} & 0.371 \\
        & Qwen3.5-27B~\cite{alibaba2026qwen35}              & 2026.02 & 60.42 & 92.83 & 34.91 & 65.09 & 529 & 1.29 & 1.13 & 0.347 \\
        & Qwen3.5-122B~\cite{alibaba2026qwen35}             & 2026.02 & 61.08 & 92.97 & 34.31 & 65.69 & 553 & 1.26 & 1.09 & 0.380 \\
        & Qwen3.5-397B~\cite{alibaba2026qwen35}             & 2026.02 & \rankthird{61.86} & 93.28 & \rankthird{33.68} & \rankthird{66.32} & 409 & 1.69 & 1.47 & \rankthird{0.326} \\
        \bottomrule
    \end{tabular}
\end{table*}

\cref{tab:main_results} presents the full evaluation results across all caption models.
We highlight several key observations:

\noindent\textbf{Observation 1: Dense QA exposes substantial cross-model gaps in factual coverage.}
Under the current construction pipeline (Gemini-assisted metadata; Gemini- and GPT-generated QA; \cref{sec:limitations}), Gemini-3.1-Pro achieves the highest Overall Accuracy (68.72\%) and Coverage (73.68\%), outperforming the second-best model Qwen3.7-Plus.
Among open-weight models, Qwen3.5-397B (61.86\%) and Qwen3-VL-235B (58.44\%) form a competitive second tier, but all models trail Gemini-3.1-Pro by a clear margin under dense probing.
These gaps are invisible to coarse-grained evaluation: a caption can appear fluent while failing on dozens of region-aligned facts probed by \method.

\noindent\textbf{Observation 2: Effective Accuracy and Coverage form a judge-dependent decomposition of Overall Accuracy.}
Across all 13 models, Effective Accuracy remains tightly clustered, indicating that when the judge commits to an answer, those answers are usually correct under our protocol.
The primary differentiator is therefore \emph{Coverage} (how often the judge answers) rather than Effective Accuracy alone.
GPT-4o exemplifies a conservative pattern under the default judge: it achieves the highest Effective Accuracy (94.46\%) but the lowest Coverage (36.43\%) and highest Uncertain Ratio (63.57\%), consistent with frequent abstention.
In contrast, Gemini-3.1-Pro maintains both high Coverage (73.68\%) and strong Overall Accuracy (68.72\%) under the same reader, corresponding to more questions resolved as supported by its captions.
Per-QA-category analysis (\cref{fig:qa_category_bars}) further shows that the Coverage gap is not uniform: all models struggle most on \emph{Count}, \emph{Camera Features}, and \emph{Attributes}, while \emph{Scene Perception} and \emph{Recognition} remain relatively strong.

\noindent\textbf{Observation 3: Longer captions do not guarantee better information density.}
Gemini-3.1-Pro produces the longest captions among proprietary models (561 tokens/image) yet ranks near the bottom on both Mean Density (1.34\textperthousand) and Global Density (1.19\textperthousand), while GPT-5.5 generates substantially shorter captions (222 tokens/image) with top-tier densities (Mean 2.53\textperthousand, Global 2.08\textperthousand).
As formalized in~\cref{sec:efficiency}, these two scores are distinct: Mean is an equal-weight average of per-image densities, whereas Global is the $M_iT_i$-weighted average; they coincide only if every image has the same request-token mass.
Across all models in~\cref{tab:main_results}, Mean Density exceeds Global Density, indicating that higher request-token mass tends to fall on lower-density images.
A similar length--density pattern appears among open-weight models: Qwen3.5-122B averages 553 tokens/image but only 1.26/1.09\textperthousand{} Mean/Global Density, whereas more compact models such as Kimi-K2.5 achieve higher density with fewer tokens.
This confirms that verbosity alone does not translate into informational value under dense QA evaluation.
% ======================================================================
\subsection{Analysis}\label{sec:analysis}

We conduct further analyses to demonstrate the diagnostic value of \method's dense, multi-dimensional evaluation.

\noindent\textbf{Domain coverage visualization.}
\cref{fig:domain_coverage} visualizes the hierarchical domain distribution of \method across 37 L1 domains and 219 L2 sub-domains.

\noindent\textbf{Per-domain breakdown.}
\cref{fig:domain_l1_radar} in~\cref{sec:appendix_domain_results} reports Overall Accuracy across all 37 L1 domains for representative models.
Per-domain accuracy is computed by pooling all QA pairs whose parent image is labeled with that L1 domain; with few images per domain, these scores are descriptive rather than stable domain-level estimates, and we do not draw conclusions at the L2 level.
Under this pooling, domains with salient global structure (\eg, \emph{Universe}, \emph{Natural Landscapes}, \emph{Pets \& Animals}) tend to score higher than fine-grained or knowledge-heavy ones such as \emph{Books}, \emph{Internet \& Telecom}, and \emph{Health \& Medical Care}.

\begin{figure}[t]
    \centering
    \includegraphics[width=\columnwidth]{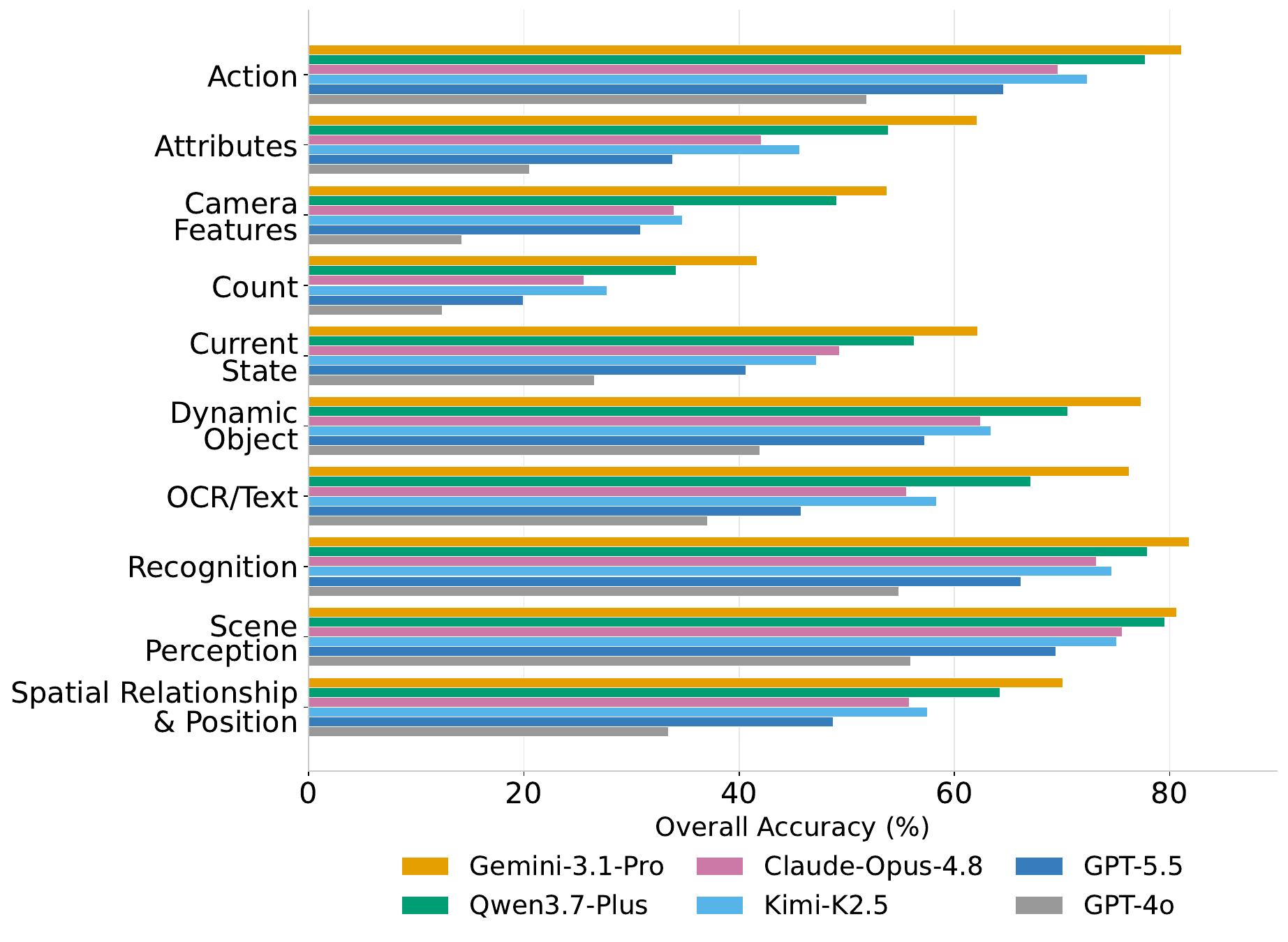}
    \caption{Per-QA-category Overall Accuracy (\%) for representative caption models (one per family).}
    \label{fig:qa_category_bars}
\end{figure}

\noindent\textbf{Per-QA-category breakdown.}
\cref{fig:qa_category_bars} visualizes Overall Accuracy across 10 QA categories for one representative model per family.
Gemini-3.1-Pro leads on nearly every axis, while GPT-4o shows the smallest profile.
All models perform relatively poorly on \emph{Count}, \emph{Camera Features}, and \emph{Attributes}, but better on \emph{Scene Perception} and \emph{Recognition}, suggesting that current VLMs still struggle with quantitative, viewpoint, and fine-grained attribute probing in detailed descriptions.

\begin{figure}[t]
    \centering
    \includegraphics[width=\columnwidth]{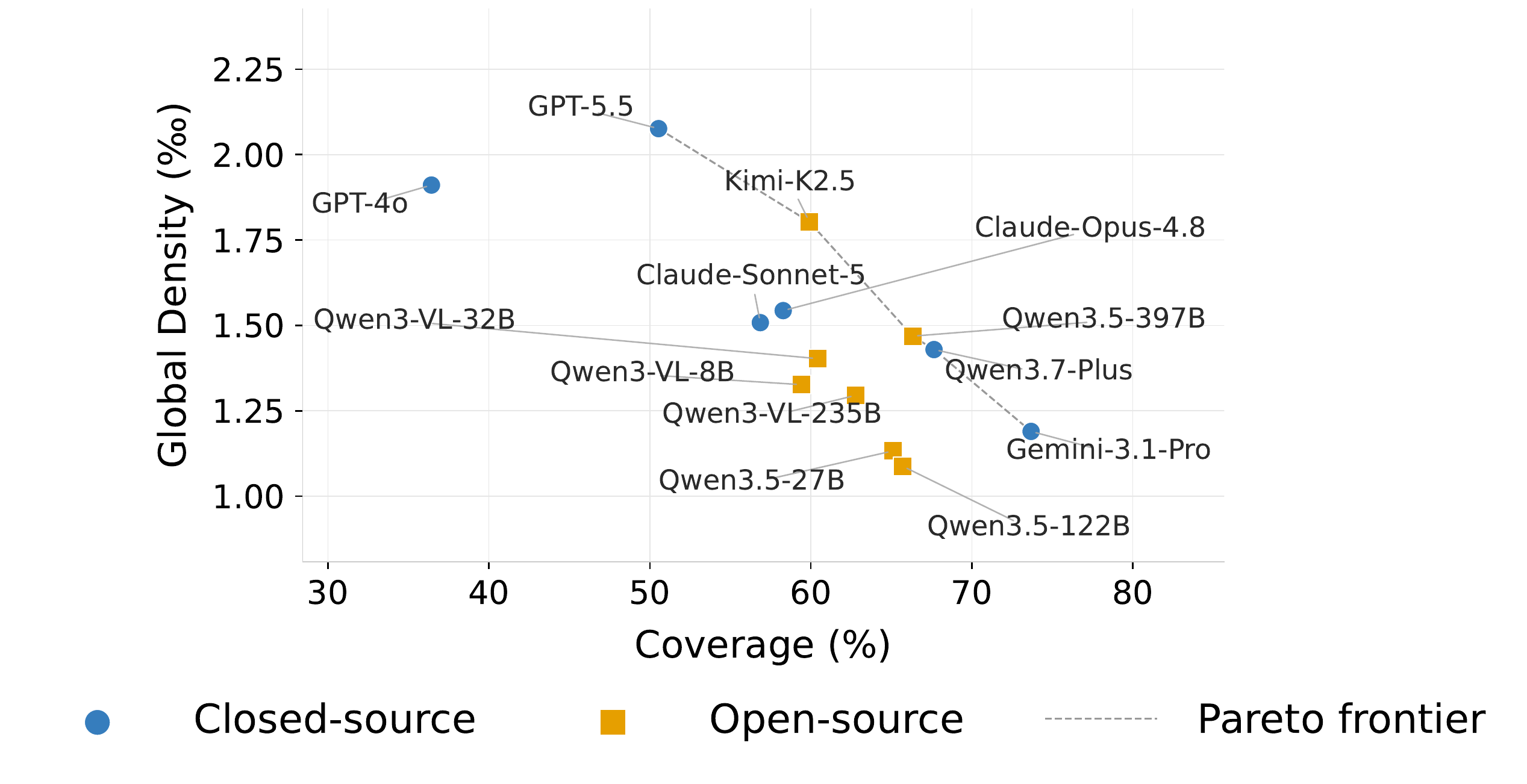}
    \caption{Coverage vs.\ Global Density for each caption model.}
    \label{fig:density_coverage}
\end{figure}

\noindent\textbf{Density vs.\ coverage trade-off.}
\cref{fig:density_coverage} visualizes the trade-off between Coverage and Global Density for each model.
Based on \cref{tab:main_results}, GPT-5.5 and GPT-4o occupy the high-density region (1.91--2.08\textperthousand) but sacrifice Coverage (36--51\%), whereas Gemini-3.1-Pro achieves the highest Coverage (73.68\%) at the cost of lower Global Density (1.19\textperthousand).
No single model simultaneously dominates both axes, revealing a fundamental trade-off between comprehensiveness and token efficiency.
The dashed Pareto frontier highlights models that are not dominated on both metrics simultaneously, including Gemini-3.1-Pro, Qwen3.7-Plus, Qwen3.5-397B, Kimi-K2.5, and GPT-5.5.

\noindent\textbf{Qualitative case studies.}
We sample one image randomly, shown in \cref{fig:qualitative_sample} in~\cref{sec:appendix_qualitative}, with captions from Gemini-3.1-Pro, Qwen3.7-Plus, and GPT-5.5, together with three probes spanning \emph{Attributes}, \emph{Recognition}, and \emph{Camera Features}.
Even on a single scene, the captions differ substantially in length and detail, and the three category probes illustrate how \method checks fine-grained visual facts that sparse or overlap-based metrics would leave unverified.

% ======================================================================
\subsection{Judge Reliability Analysis}\label{sec:judge_reliability}

Stage~2 evaluation relies on a language judge to answer MCQs from captions alone.
We re-evaluate the top-3 caption models from \cref{tab:main_results}---Gemini-3.1-Pro, Qwen3.7-Plus, and Qwen3.5-397B---with three judges spanning different families and scales: Qwen3-32B, Qwen3-8B~\cite{yang2025qwen3}, and Llama-3.1-8B-Instruct~\cite{grattafiori2024llama3}, all using the same Stage~2 prompt and QA set.

\cref{tab:judge_reliability} reports Overall Accuracy under each judge.
Absolute scores vary substantially---Llama-3.1-8B-Instruct scores highest, while the Qwen3 judges yield lower, more compressed values---but the ranking is unchanged: Gemini-3.1-Pro $>$ Qwen3.7-Plus $>$ Qwen3.5-397B.
We attribute the absolute-score gap mainly to differences in how aggressively each judge selects option~E versus committing to A--D:
judges that refuse less often can inflate Overall Accuracy, because forcing a choice among A--D yields a non-zero chance of matching the ground truth even when the caption is incomplete, whereas selecting E always counts as incorrect under Overall Accuracy.
Thus \method reduces open-ended scorer bias by tying correctness to MCQ answers, yet absolute Overall Accuracy remains strongly \emph{reader-dependent}; it is reliable for \emph{comparative} evaluation under a fixed judge and can be obtained with inexpensive open-weight readers, unlike LLM-as-scorer protocols that typically require strong proprietary models.

\begin{table}[t]
    \centering
    \caption{Judge reliability on top-3 caption models. Overall Accuracy (\%) under three language judges.}
    \label{tab:judge_reliability}
    \small
    \setlength{\tabcolsep}{4pt}
    \begin{tabular}{lccc}
        \toprule
        Caption Model & Qwen3-32B & Qwen3-8B & Llama-3.1-8B \\
        \midrule
        Gemini-3.1-Pro~\cite{google2026gemini31} & \rankfirst{68.72} & \rankfirst{43.26} & \rankfirst{78.16} \\
        Qwen3.7-Plus~\cite{alibaba2026qwen37plus}   & \ranksecond{63.07} & \ranksecond{38.53} & \ranksecond{74.27} \\
        Qwen3.5-397B~\cite{alibaba2026qwen35}       & \rankthird{61.86} & \rankthird{38.06} & \rankthird{73.07} \\
        \bottomrule
    \end{tabular}
\end{table}

%% file: sec/6_conclusion.tex
\section{Conclusion}\label{sec:conclusion}

We presented \method, a full-scene dense QA benchmark for detailed image caption evaluation, with full-scene coverage as a design goal over coarse semantic regions.
Following the principle that dense caption supervision calls for equally dense assessment, \method anchors 74 QA pairs per image to retained semantic regions and probes 10 categories across 37 L1 / 219 L2 domains.
Captions are scored by a three-stage protocol---generation, caption-only multiple-choice answering with an Uncertain option, and multi-dimensional metric computation---which converts unconstrained scalar scoring into structured MCQ reading, thereby reducing open-ended scorer bias while remaining reader-dependent, and supports reproducible comparative ranking under a fixed judge.
Competency metrics (Overall/Effective Accuracy, Uncertain Ratio, Coverage) decompose Overall Accuracy into a judge-dependent answering rate and committed-answer accuracy, while efficiency metrics (Global Density, Density CV) expose when longer captions fail to carry more verifiable information.
Across 13 VLMs, \method reveals substantial Coverage gaps between comprehensive and conservative captioners and a persistent competency--efficiency trade-off that coarse evaluation often conceals.
We discuss limitations of \method in~\cref{sec:limitations}.

%% file: sec/7_ref.tex
\clearpage
\small
\bibliographystyle{ieeenat_fullname}
\bibliography{main}

%% file: sec/8_appendix.tex
\newpage
\appendix

\section{Limitations}\label{sec:limitations}

We highlight several limitations of \method.
\textbf{First}, absolute Overall Accuracy depends on the Stage~2 language judge: judges differ in how readily they select the Uncertain option versus guessing among A--D, which can shift absolute scores even when caption rankings remain stable.
Effective Accuracy and Coverage likewise provide only a judge-dependent proxy for unanswered versus incorrectly resolved probes---not a validated causal separation of caption omission from hallucination---so they should be read relative to a fixed reader.
Our protocol therefore reduces open-ended LLM-as-scorer bias via structured MCQ reading, but does \emph{not} eliminate reader-model dependence; we recommend reporting results under a fixed judge (or sharing judge outputs) and treating absolute scores as judge-conditioned.
\textbf{Second}, region metadata is produced by Gemini-3.1-Pro, and MCQs are generated by Gemini-3.1-Pro and GPT-5.5 before human fact-checking.
Human QC targets image-grounded correctness under the protocol in~\cref{sec:qa_qc}, but does not establish that the benchmark is neutral across model families: wording, fact granularity, category emphasis, and distractor style may still align with the generators' captioning habits.
Both generators are also among the evaluated caption models.
We therefore interpret strong Gemini-3.1-Pro Overall Accuracy and Coverage as results under the current Gemini-assisted construction pipeline (Gemini metadata; Gemini/GPT QA), and leave cross-generator validation---e.g., independent non-Gemini QA subsets---to future work.
\textbf{Third}, region segmentation with YOLOv26-seg and SAM3 targets coarse full-scene coverage of discrete objects, parts, and background elements, but we do not report region recall against human-enumerated salient entities; complex scenes may still omit important objects or stuff, so ``full-scene'' remains a design goal and the benchmark is densely probed over retained regions rather than exhaustively instance-complete.
\textbf{Fourth}, with 346 images spanning 219 L2 sub-domains (often only 1--2 images each), the benchmark prioritizes broad taxonomy coverage and QA density over statistical representativeness or stable per-domain estimation; future extensions could enlarge the image pool while preserving the region-aligned dense QA protocol.
\textbf{Fifth}, QA verification is partitioned among three annotators with expert adjudication on all edits and a 5\% sample of accepts; we do not report full-set inter-annotator agreement or accept/edit/delete rate breakdowns, so residual label noise and revise-heavy items cannot be ruled out from deletion counts alone.

\section{Implementation Details}\label{sec:appendix_prompts}

In this section, we provide the complete prompts used in the \method construction and evaluation pipeline.

% ======================================================================
\subsection{Metadata Generation Prompt}\label{sec:appendix_meta_prompt}

The following prompt in Figure~\ref{fig:meta_prompt} is used to generate structured metadata for each segmented region via Gemini-3.1-Pro~\cite{google2026gemini31}.
The prompt instructs the model to analyze both the original image (for spatial context) and the masked region (for fine-grained attributes), and to output a structured JSON object.
The schema is designed to handle both discrete objects (``things'') and background/environmental regions (``stuff''), supporting broad scene coverage.

\vspace{2mm}
\begin{figure*}[t]
\begin{minipage}[t]{0.48\textwidth}
\begin{lstlisting}[basicstyle=\ttfamily\scriptsize, breaklines=true, frame=single, framesep=3pt, xleftmargin=2pt, xrightmargin=2pt, columns=fullflexible]
Role: You are an advanced Visual Perception & Scene Analysis Engine specialized in fine-grained semantic auditing of image segments.
Task: Extract high-precision Metadata (Meta Info) for the target masked segment. The target may be a discrete object ("Thing") or an environmental/structural element ("Stuff").

Inputs:
- Image 1 (Context): The complete original image used for spatial localization, global context, and environmental conditions.
- Image 2 (Masked): The high-resolution view of the specific target segment to be analyzed.

Strict JSON Requirements:
1. Conditional Fields: If a field does not apply to the target (e.g., 'text_and_markings' for an empty sky, or 'interaction_targets' for a road), return `null` or an empty array `[]`. Do not hallucinate.
2. Coordinates: Provide 'bounding_box' in [ymin, xmin, ymax, xmax] format, normalized (0-1000) based on Image 1.
3. Element & Component Audit: For objects, list sub-parts (e.g., wheels). For background regions, list constituent elements (e.g., tiles on a roof, patches on a road).

Schema Structure:
{
  "spatial_and_localization": {
    "bounding_box": [ymin, xmin, ymax, xmax],
    "global_position": "string (e.g., 'center foreground', 'upper background')",
    "spatial_layer": "string (e.g., 'foreground', 'midground', 'background')",
    "visibility_state": {
      "is_occluded": boolean,
      "occluded_by": "string or null",
      "visible_range": "string (e.g., 'fully visible', 'partially cut off by frame')"
    }
  },
  "identity_and_taxonomy": {
    "segment_nature": "string ('thing' for discrete objects/instances, 'stuff' for background/environmental regions)",
    "broad_category": "string (e.g., 'Vehicle', 'Vegetation', 'Sky', 'Architecture')",
    "specific_type": "string (e.g., 'Sedan', 'Oak Tree', 'Overcast Sky', 'Brick Wall')",
    "primary_function_or_role": "string (e.g., 'transportation', 'environmental background', 'pedestrian walkway')"
  },
\end{lstlisting}
\end{minipage}\hfill
\begin{minipage}[t]{0.48\textwidth}
\begin{lstlisting}[basicstyle=\ttfamily\scriptsize, breaklines=true, frame=single, framesep=3pt, xleftmargin=2pt, xrightmargin=2pt, columns=fullflexible]
  "detailed_appearance": {
    "key_identifying_feature": "string (A one-sentence unique description of this segment)",
    "overall_color_palette": ["string"],
    "materials_and_textures": ["string (e.g., 'smooth concrete', 'rough fabric', 'reflective glass')"],
    "sub_components_or_elements": [
      {
        "element_name": "string (e.g., 'front bumper' for car, or 'cloud patches' for sky)",
        "description": "string",
        "quantity": number
      }
    ],
    "text_and_markings": [
      {
        "content": "string (Exact transcribed text/OCR)",
        "type": "string (e.g., 'Brand Logo', 'Graffiti', 'Street Sign Text')",
        "placement": "string",
        "is_partially_obscured": boolean
      }
    ]
  },
  "dynamics_and_interactions": {
    "motion_or_physical_state": "string (e.g., 'static', 'moving rapidly', 'flowing liquid')",
    "interaction_details": "string or null (For objects: what it interacts with. For surfaces/backgrounds: what objects are currently occupying or traversing it, e.g., 'cars driving on this road')"
  },
  "global_scene_context": {
    "lighting_and_weather": "string (e.g., 'harsh mid-day sunlight with sharp shadows', 'overcast and rainy')",
    "time_of_day": "string (e.g., 'Golden hour', 'Night')",
    "overall_atmosphere": "string (e.g., 'Busy urban rush', 'Serene nature')"
  }
}

Instructions:
- Segment Adaptability: First identify if the segment is a 'thing' or 'stuff' in `segment_nature`, then tailor the granularity accordingly.
- Quantitative Precision: Count specific visible sub-elements or features where applicable.
- Absolute OCR: Transcribe every letter or symbol found within the masked area. If none exist, return an empty array.
- Output ONLY the raw JSON string. Do not include markdown code blocks.
\end{lstlisting}
\end{minipage}
\caption{Complete prompt for structured metadata generation per region.}
\label{fig:meta_prompt}
\end{figure*}

% ======================================================================
\subsection{Dense QA Generation Prompt}\label{sec:appendix_qa_prompt}

The prompt in Figure~\ref{fig:qa_prompt} is used by both Gemini-3.1-Pro and GPT-5.5 to generate multiple-choice questions for each segmented region.
The prompt emphasizes three key design principles:
(1)~\emph{Zero ambiguity}---every question must include a unique identifying description based on the region's global position and key features;
(2)~\emph{Natural language}---no technical jargon such as ``masked region'' or ``target area'';
(3)~\emph{Exhaustive coverage}---questions must span all 10 QA categories.

\vspace{2mm}
\begin{figure*}[t]
\begin{minipage}[t]{0.48\textwidth}
\begin{lstlisting}[basicstyle=\ttfamily\scriptsize, breaklines=true, frame=single, framesep=3pt, xleftmargin=2pt, xrightmargin=2pt, columns=fullflexible]
You are a professional dataset annotator. Your task is to generate high-quality multiple-choice questions (MCQs) to evaluate the accuracy of detailed image captions.

Input:
1. Original Image: The full scene context.
2. Masked Image: Used ONLY for your reference to identify which region to focus on.
3. Meta Info: {META_INFO} (Contains the ground-truth attributes and precise global position).

Task:
Generate the MAXIMUM possible number of QA pairs. These questions will be used to verify if a caption correctly describes the specific region.

Requirements:

1. Absolute Zero Ambiguity (Critical)
- You MUST assume the Original Image contains multiple similar objects (e.g., multiple dogs, multiple people).
- Every 'question' MUST start with or include a unique identifying description based on the 'global_position' and 'key_identifying_feature' from Meta Info.
- NEVER use generic terms like "the dog," "the animal," or "the person" alone.
- MANDATORY FORMAT: Use "The [Category] [Spatial Position/Context] [Unique Feature]...".
  * Bad: "What color is the dog's harness?"
  * Good: "What color is the harness worn by the English Bulldog positioned between the orange trash bin and the woman in yellow?"

2. "Natural Look" Constraint (No Technical Jargon)
- NEVER mention "masked image," "highlighted region," "target," "masked area," or "bounding box" in the questions or options.
- The questions must feel like they are asked by a human looking at the ORIGINAL image only.
- Avoid phrases like "in this region" or "of the identified target."

3. Coverage Requirement (10 Categories)
Exhaustively cover the target's details using these categories:
A. Recognition: Specific breed/type (e.g., English Bulldog, not just dog).
B. Attributes: Details (head wrinkles, harness color, paw markings).
C. OCR/Text: Any text/logos on the object itself.
D. Scene Perception: Environment type, background, lighting, indoor/outdoor, weather, materials, or the general setting (e.g., urban sidewalk) etc.
E. Count: Count features ON the object (e.g., "How many white patches are on the neck of the Bulldog..."), or Approximate number of visible items.
\end{lstlisting}
\end{minipage}\hfill
\begin{minipage}[t]{0.48\textwidth}
\begin{lstlisting}[basicstyle=\ttfamily\scriptsize, breaklines=true, frame=single, framesep=3pt, xleftmargin=2pt, xrightmargin=2pt, columns=fullflexible]
F. Current State: Condition (well-groomed, healthy, open/closed, on/off, full/empty).
G. Dynamic Object: Any motion blur or moving objects on this specific object.
H. Action: Details (e.g., standing on all fours, tongue visible).
I. Camera Features: Viewpoint/angle relative to this specific object (e.g., "From what angle is the silver-harnessed Bulldog captured?").
J. Spatial Relationship & Position: Specifically how the object is situated relative to other anchors (e.g., "Is the Bulldog closer to the orange bin?").

4. Style & Output
- 4 options (A, B, C, D), one correct answer.
- Randomize the Position of the Correct Answer: Do NOT always place the correct answer in option A. You MUST randomly distribute the correct answer among A, B, C, and D for each question.
- Distractors must be plausible based on the scene but factually wrong for this object.
- Aim for 15-20+ QA pairs.

Output Format (JSON):
{
  "object_id": "{OBJECT_ID}",
  "qas": [
    {
      "category": "... one from the 10 categories (A-J) ...",
      "question": "[Unambiguous Description of Object], [Question Topic]?",
      "options": [
        "A. Option text",
        "B. Option text",
        "C. Option text",
        "D. Option text"
      ],
      "correct_answer": "A"  // The label (A, B, C, or D) of the correct option
    }
  ]
}
\end{lstlisting}
\end{minipage}
\caption{Complete prompt for dense QA generation per region.}
\label{fig:qa_prompt}
\end{figure*}

% ======================================================================
\subsection{Evaluation Judge Prompt}\label{sec:appendix_judge_prompt}

The following prompt in Figure~\ref{fig:judge_prompt} is used during Stage~2 of the evaluation protocol, where a language judge answers questions based solely on the generated caption.
The prompt explicitly instructs the judge to use \textbf{only} the information within the caption tags and to select option~E (``Not mentioned / Cannot be determined'') when the caption lacks sufficient information, enabling the computation of Effective Accuracy.

\vspace{2mm}
\noindent\begin{minipage}{\columnwidth}
\begin{lstlisting}[basicstyle=\ttfamily\scriptsize, breaklines=true, frame=single, framesep=3pt, xleftmargin=2pt, xrightmargin=2pt, columns=fullflexible]
You are provided with a text description of an image labeled as <CAPTION>. Your task is to answer the question using **ONLY** the information contained within the <CAPTION> tags. Ignore all outside knowledge and do not infer details not explicitly mentioned.

<CAPTION>
{CAPTION_HERE}
</CAPTION>

QUESTION:
{QUESTION_HERE}

OPTIONS:
A. {OPTION_A}
B. {OPTION_B}
C. {OPTION_C}
D. {OPTION_D}
E. Not mentioned / Cannot be determined

Based solely on the <CAPTION> above, respond with only the letter (A, B, C, D or E).
\end{lstlisting}
\end{minipage}
\vspace{1mm}
\captionof{figure}{Complete prompt for caption-based QA answering (Stage~2). The judge selects from options A--E, where E indicates insufficient information in the caption.}
\label{fig:judge_prompt}

\input{sec/appendix_domain_taxonomy}

\section{Per-Domain Accuracy Results}\label{sec:appendix_domain_results}

\begin{figure*}[t]
    \centering
    \includegraphics[width=0.92\textwidth]{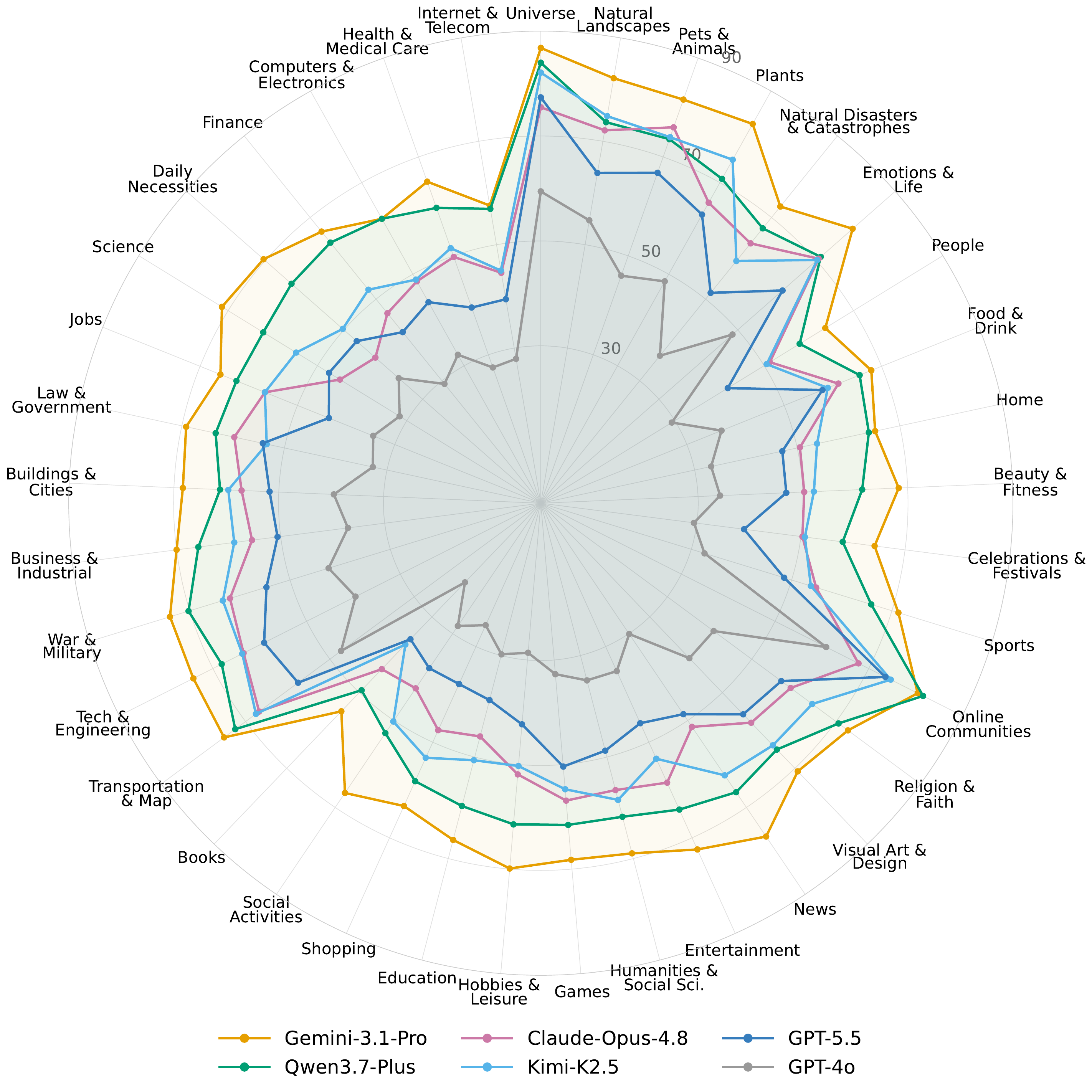}
    \caption{Per-L1-domain Overall Accuracy (\%) for six representative caption models across all 37 L1 domains.}
    \label{fig:domain_l1_radar}
\end{figure*}

\cref{fig:domain_l1_radar} presents Overall Accuracy (\%) of six representative caption models across all 37 L1 domains.
Per-domain accuracy is obtained by pooling all QA pairs whose parent image is labeled with the corresponding L1 domain; given the limited images per domain, these values are descriptive taxonomy-level snapshots rather than statistically stable domain estimates.

\section{Qualitative Sample}\label{sec:appendix_qualitative}

\begin{figure*}[t]
    \centering
    \includegraphics[width=\textwidth]{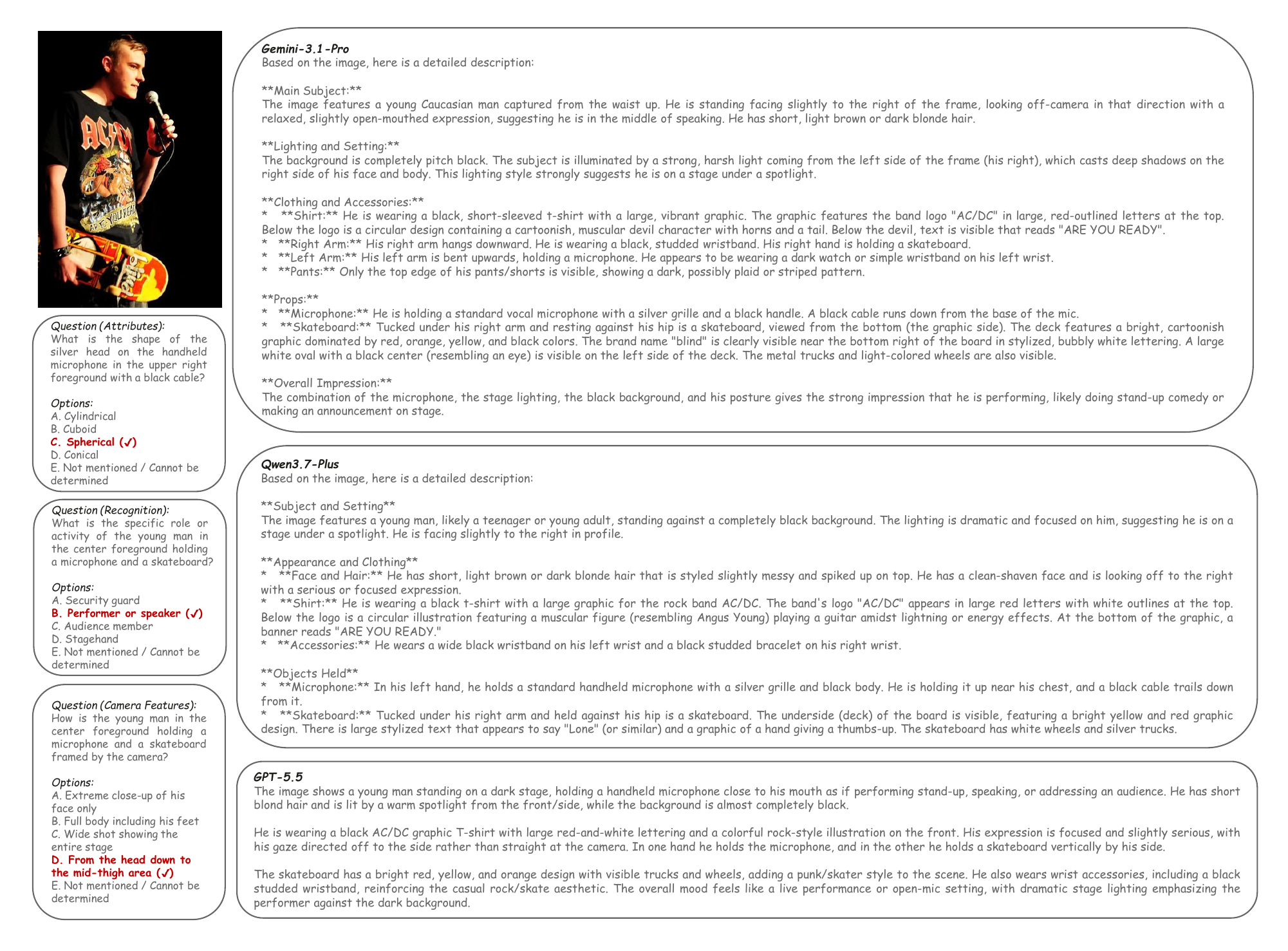}
    \caption{Qualitative sample on \method: one image, captions from three models, and three QA probes
    (\emph{Attributes}, \emph{Recognition}, and \emph{Camera Features}).}
    \label{fig:qualitative_sample}
\end{figure*}

\cref{fig:qualitative_sample} provides a qualitative illustration of \method on one image.
We show captions generated by Gemini-3.1-Pro, Qwen3.7-Plus, and GPT-5.5, along with three representative QA pairs covering \emph{Attributes}, \emph{Recognition}, and \emph{Camera Features}.

%% file: sec/appendix_domain_taxonomy.tex
\section{Domain Taxonomy}\label{sec:appendix_domain}

This section lists the complete two-level domain taxonomy used for hierarchical domain-guided image selection, comprising 37 L1 domains and 219 L2 sub-domains.

\begin{table*}[t]
\centering
\caption{Complete L1/L2 domain taxonomy (37 L1 domains, 219 L2 sub-domains).}
\label{tab:domain_taxonomy}
\footnotesize
\setlength{\tabcolsep}{4pt}
\renewcommand{\arraystretch}{1.15}
\begin{tabular}{@{}p{0.20\textwidth}p{0.77\textwidth}@{}}
\toprule
\textbf{L1 Domain} & \textbf{L2 Sub-domains} \\
\midrule
    Beauty \& Fitness & Beauty Services \& Spas; Cosmetology \& Beauty Professionals; Face \& Body Care; Fashion \& Style; Fitness; Hair Care \\
    Books & Art Books; Economy and Management; Education Books; Life Books; Literature Books; Natural Science Popularization Books; Social Sciences \\
    Buildings \& Cities & Administrative Office; Countryside \& Farmland; Facilities; Historical Buildings; Hotels \& Accommodations; House \& Apartment; Modern Architecture; Property Development; Restaurants; Urban Landscapes \\
    Business \& Industrial & Advertising \& Marketing; Agriculture \& Forestry; Automotive Industry; Business Operations; Business People; Business Services; Chemicals Industry; Construction \& Maintenance; Energy \& Utilities; Hospitality Industry; Industrial Materials \& Equipment; Manufacturing; Metals \& Mining; Pharmaceuticals \& Biotech; Retail Trade; Shipping \& Logistics \\
    Celebrations \& Festivals & Birthdays \& Anniversaries; Cultural Festivals; New Year Celebrations; Traditional Festivals; Weddings \& Celebrations \\
    Computers \& Electronics & CAD \& CAM; Computer Hardware; Consumer Electronics; Electronics \& Electrical; Enterprise Technology; Programming \\
    Daily Necessities & Apparel; Home Goods; Hygiene and Personal Care \\
    Education & Campus; Classroom; Lecture; Online Education; Other; Tutorial; Workshop \\
    Emotions \& Life & Children \& Teens; Everyday Moments; Family Life; Love \& Friendship \\
    Entertainment & Movies; Music \& Audio; Offbeat; Performing Arts; Stand-up Comedy; TV \& Video; musical instrument \\
    Finance & Banking; Financial Planning \& Management; Money \\
    Food \& Drink & Drinks; Food; Snack \\
    Games & Arcade \& Coin-Op Games; Board Games; Computer \& Video Games; Dice Games; Family-Oriented Games \& Activities; Table Games \\
    Health \& Medical care & Aging \& Geriatrics; Health Conditions; Health Education \& Medical Training; Health Foundations \& Medical Research; Medical Devices \& Equipment; Medical Facilities \& Services; Medical Imaging; Nursing; Oral \& Dental Care; Pediatrics; Pharmacy; Public Health; Reproductive Health; Vision Care; Women's Health; paramedics \\
    Hobbies \& Leisure & Clubs \& Organizations; Crafts; Merit Prizes \& Contests; Outdoors; Radio Control \& Modeling; Recreational Aviation; Special Occasions; Sweepstakes \& Promotional Giveaways; Water Activities \\
    Home & Bed \& Bath; HVAC \& Climate Control; Home \& Interior Decor; Home Appliances; Home Furnishings; Home Improvement; Home Storage \& Shelving; Kitchen \& Dining; Laundry; Patio, Lawn \& Garden \\
    Humanities \& Social Sci. & History; Sociology \\
    Internet \& Telecom & Communications Equipment; Email \& Messaging; Mobile \& Wireless; Service Providers; Teleconferencing; Web Services \\
    Jobs & Factories \& Manufacturing; Office Work; Outdoor Work; Professionals; Work Environment \\
    Law \& Government & Government; Legal; Public Safety \\
    Natural Disasters \& Catastrophes & Earthquakes; Fires; Floods; Storms \& Hurricanes \\
    Natural Landscapes & Forests \& Vegetation; Mountains \& Rivers; Oceans \& Beaches; Weather \& Sky \\
    News & Health News; Local News; Technology News \\
    Online Communities & Photo \& Video Sharing \\
    People & Culturally Significant People; Group Portraits; Portraits; Professional People \\
    Pets \& Animals & Animal Products \& Services; Birds \& Flying Creatures; Insects \& Small Animals; Pets; Wild Animals; fish \\
    Plants & Natural Plants; domestic plants \\
    Religion \& Faith & Believers \& Faith; Religious Buildings; Religious Ceremonies; Religious Symbols \\
    Science & Biology; Chemistry; Geography; Physics \\
    Shopping & Antiques \& Collectibles; Apparel; Consumer Resources; Mass Merchants \& Department Stores; Swap Meets \& Outdoor Markets; Toys \\
    Social Activities & Crowds on the streets; Cultural and recreational activities; Meetings for Exchange; Political activity \\
    Sports & Animal Sports; Ball Sports; Combat Sports; Extreme Sports; Individual Sports; International Sports Competitions; Motor Sports; Sport Scores \& Statistics; Sporting Goods; Sports Coaching \& Training; Sports Fan Gear \& Apparel; Team Sports; Water Sports; Winter Sports; track and field \\
    Tech \& Engineering & Agriculture; Architecture Eng.; Electronics; Energy Power; Materials; Mechanical Eng. \\
    Transportation \& Map & Air traffic; Land transportation; Road conditions; Traffic Signs; Underground traffic; Water transportation \\
    Universe & Near-Earth Space; Outer Space \\
    Visual Art \& Design & Arts \& Crafts; Comics \& Animation; Painting; Sculpture \& Installation Art; cover; poster \\
    War \& Military & Battle Scenes; Firearms \& Weapons; Logistics; Military \& Soldiers; Military Equipment; War \& Conflict; War Relics; military site \\
\bottomrule
\end{tabular}
\end{table*}